\documentclass[10pt,journal,compsoc]{IEEEtran}

\PassOptionsToPackage{table}{xcolor}
\def\RevisionPreambleLight{1}
\ifdefined\pdfminorversion\pdfminorversion=4\fi
\ifdefined\pdfobjcompresslevel\pdfobjcompresslevel=0\fi

\makeatletter
\PassOptionsToPackage{table}{xcolor}
\@ifpackageloaded{xcolor}{}{\usepackage{xcolor}}
\@ifpackageloaded{soul}{}{\usepackage{soul}}
\@ifpackageloaded{pifont}{}{\usepackage{pifont}}
\@ifpackageloaded{tikz}{}{\usepackage{tikz}}
\makeatother
\definecolor{revisionHighlightColor}{RGB}{0,0,255}
\providecommand{\blue}[1]{}
\renewcommand{\blue}[1]{\textcolor{revisionHighlightColor}{#1}}
\providecommand{\RegisterRevisionHighlightCommands}{%
  \ifdefined\cite\soulregister\cite7\fi
  \ifdefined\ref\soulregister\ref7\fi
  \ifdefined\cref\soulregister\cref7\fi
  \ifdefined\Cref\soulregister\Cref7\fi
  \ifdefined\eqref\soulregister\eqref7\fi
  \ifdefined\textbf\soulregister\textbf7\fi
  \ifdefined\textit\soulregister\textit7\fi
  \ifdefined\emph\soulregister\emph7\fi
}
\RegisterRevisionHighlightCommands
\makeatletter
\@ifundefined{hlmaskenv}{%
  \newsavebox{\hlmasksavebox}%
}{}
\makeatother

\ifdefined\RevisionPreambleLight
  \def\RevisionPreambleMaybeEnd{ }
\else
  \let\RevisionPreambleMaybeEnd\relax
\fi
\RevisionPreambleMaybeEnd

\usepackage{mathtools,bm,amsmath,amssymb,amsfonts,amsthm,nicefrac,latexsym,dsfont,extarrows,bbm}
\usepackage{cases,subeqnarray}
\usepackage{algorithm,algorithmicx,algpseudocode}
\usepackage{booktabs,multirow,subcaption,graphicx,epsfig,epstopdf,float,stfloats,bigstrut,makecell,array}
\usepackage[most]{tcolorbox}
\usepackage{xcolor,color,colortbl,textcomp,pifont,microtype}
\usepackage{multicol,balance,flushend,setspace,titletoc}
\usepackage[noadjust]{cite}
\usepackage{url,xr-hyper}
\makeatletter
\@ifpackageloaded{wasysym}{}{\usepackage[nointegrals]{wasysym}}
\makeatother

\definecolor{citecolor}{HTML}{0071bc}
\ifdefined\RevisionSupplementColorLinks
\usepackage[pagebackref=false,breaklinks=true,colorlinks=true,bookmarks=true,linkcolor=black,citecolor=citecolor,urlcolor=citecolor,filecolor=citecolor]{hyperref}
\else
\usepackage[pagebackref=false,breaklinks=true,colorlinks=false,pdfborder={0 0 0},bookmarks=true,linkbordercolor=white,citebordercolor=white,urlbordercolor=white,filebordercolor=white]{hyperref}
\fi
\makeatletter
\@ifundefined{Hy@setpdfversionfalse}{}{\Hy@setpdfversionfalse}
\makeatother
\usepackage[nameinlink]{cleveref}
\RegisterRevisionHighlightCommands

\usepackage{tikz}
\definecolor{lime}{HTML}{A6CE39}
\DeclareRobustCommand{\orcidicon}{\begin{tikzpicture}\draw[lime, fill=lime] (0,0) circle [radius=0.16] node[white] {{\fontfamily{qag}\selectfont \tiny ID}};\draw[white, fill=white] (-0.0625,0.095) circle [radius=0.007];\end{tikzpicture}\hspace{-2mm}}
\foreach \x in {A, ..., Z}{\expandafter\xdef\csname orcid\x\endcsname{\noexpand\href{https://orcid.org/\csname orcidauthor\x\endcsname}{\noexpand\orcidicon}}}

\theoremstyle{plain}

\theoremstyle{remark}

\crefname{section}{Sec.}{Secs.}
\crefname{subsection}{Sec.}{Secs.}
\crefname{subsubsection}{Sec.}{Secs.}
\crefname{figure}{Fig.}{Figs.}
\crefname{table}{Table}{Tables}
\crefname{equation}{Eq.}{Eqs.}
\crefname{algorithm}{Alg.}{Algs.}
\crefname{assumption}{Assumption}{Assumptions}
\crefname{theorem}{Theorem}{Theorems}
\crefname{lemma}{Lemma}{Lemmas}
\crefname{proposition}{Proposition}{Propositions}
\crefname{corollary}{Corollary}{Corollaries}
\crefname{definition}{Definition}{Definitions}
\crefformat{equation}{#2Eq.~#1#3}
\crefmultiformat{equation}{#2Eqs.~#1#3}{ and~#2#1#3}{, #2#1#3}{ and~#2#1#3}
\Crefformat{equation}{#2Eq.~#1#3}
\Crefmultiformat{equation}{#2Eqs.~#1#3}{ and~#2#1#3}{, #2#1#3}{ and~#2#1#3}

\definecolor{lightkeycolor}{RGB}{255,240,240}
\definecolor{lightgray}{RGB}{234,234,234}

\def\x{{\mathbf x}}

\ifdefined\pdfminorversion\pdfminorversion=4\fi
\ifdefined\pdfinclusioncopyfonts\pdfinclusioncopyfonts=1\fi
\IfFileExists{macros.tex}{\input{macros}}{}

\let\RevisionPreambleLight\undefined

\usepackage{amsmath}
\usepackage{gensymb}
\usepackage{diagbox}
\usepackage{graphics}
\usepackage{epsfig}
\usepackage{wasysym}
\usepackage{threeparttable}
\usepackage{siunitx}
\usepackage{xspace}
\usepackage{siunitx}
\usepackage{setspace}
\usepackage{graphicx}
\usepackage{subfig}
\usepackage{placeins}
\usepackage{blindtext}
\usepackage{amssymb}
\usepackage{threeparttable}
\usepackage{color}
\usepackage[normalem]{ulem}
\usepackage{multirow}
\usepackage{float}
\usepackage{wrapfig}
\usepackage{amsfonts}
\usepackage{bm}
\usepackage{bbm}
\usepackage{array}
\usepackage{tabulary}	
\usepackage{microtype}
\usepackage{booktabs}
\usepackage{xspace}
\usepackage[table]{xcolor}
\usepackage{colortbl}
\usepackage{diagbox}
\usepackage{rotating}
\usepackage{booktabs}
\usepackage{overpic}
\usepackage{enumitem}
\usepackage{colortbl}
\usepackage{soul}
\usepackage{url}
\usepackage{makecell, verbatim, sidecap}
\usepackage{algorithm}
\usepackage{algorithmic}
\usepackage{cite}
\usepackage{tikz}
\usepackage{mathrsfs}
\usepackage{longtable}
\usepackage{multicol}
\usepackage{calc}
\usepackage{pifont}
\usepackage{fontawesome5}
\usepackage{ragged2e}

\usepackage{tikz}
\usepackage[edges]{forest}
\definecolor{hidden-draw}{RGB}{20,68,106}
\definecolor{hidden-pink}{RGB}{255,245,247}
\usepackage{placeins}

\definecolor{citecolor}{HTML}{0071bc}

\usepackage{xr-hyper}
\usepackage[pagebackref=false,breaklinks=true,colorlinks,citecolor=citecolor,linkcolor=citecolor,bookmarks=true]{hyperref}
\usepackage[nameinlink]{cleveref}
\IfFileExists{supp.aux}{\externaldocument[supp:]{supp}}{\IfFileExists{supp.aux}{\externaldocument[supp:]{supp}}{}}

\usepackage{tikz}
\definecolor{lime}{HTML}{A6CE39}
\DeclareRobustCommand{\orcidicon}{%
\begin{tikzpicture}
\draw[lime, fill=lime] (0,0) 
circle [radius=0.16] 
node[white] {{\fontfamily{qag}\selectfont \tiny ID}};\draw[white, fill=white] (-0.0625,0.095) 
circle [radius=0.007];\end{tikzpicture}
\hspace{-2mm}}
\foreach \x in {A, ..., Z}{%
\expandafter\xdef\csname orcid\x\endcsname{\noexpand\href{https://orcid.org/\csname orcidauthor\x\endcsname}{\noexpand\orcidicon}}
}

\definecolor{scorered}{HTML}{e4485a}
\definecolor{scoreblue}{HTML}{4a7ee8}
\definecolor{scoregreen}{HTML}{80ba0e}

\definecolor{purple0}{HTML}{e9e9f3}
\definecolor{purple}{HTML}{dcdaed}
\definecolor{purple1}{HTML}{bab6da}
\definecolor{blue0}{HTML}{b6d9f0}
\definecolor{blue1}{HTML}{80b1d1}
\definecolor{blue2}{HTML}{2a8ed1}
\definecolor{blue3}{HTML}{0071bc}

\definecolor{mygray00}{gray}{.3}
\definecolor{mygray0}{gray}{.6}
\definecolor{mygray}{gray}{.85}
\definecolor{mygray1}{gray}{.9}
\definecolor{mygray2}{gray}{.95}

\crefname{section}{Sec.}{Secs.}
\crefname{subsection}{Sec.}{Secs.}
\crefname{subsubsection}{Sec.}{Secs.}
\crefname{figure}{Fig.}{Figs.}
\crefname{table}{Table}{Tables}
\crefname{equation}{Eq.}{Eqs.}
\crefname{algorithm}{Alg.}{Algs.}
\crefname{theorem}{Theorem}{Theorems}
\crefname{lemma}{Lemma}{Lemmas}
\crefname{proposition}{Proposition}{Propositions}
\crefname{corollary}{Corollary}{Corollaries}
\crefname{remark}{Remark}{Remarks}
\crefname{definition}{Definition}{Definitions}

\hypersetup{
    colorlinks=true,
    linkcolor=blue,
    filecolor=blue,
    urlcolor=blue,
}

\makeatletter
\newcommand{\thickhline}{%
    \noalign {\ifnum 0=`}\fi \hrule height 1pt
    \futurelet \reserved@a \@xhline
}
\makeatother

\makeatletter
\DeclareRobustCommand\onedot{\futurelet\@let@token\@onedot}
\def\@onedot{\ifx\@let@token.\else.\null\fi\xspace}

\def\etal{\emph{et al}\onedot}

\makeatother

\newcommand{\app}{\raise.17ex\hbox{$\scriptstyle\sim$}}

\begin{document}

\title{Anomaly Detection and Generation with Diffusion Models: A Survey}

\author{
Yang Liu\orcidG{},
Jing Liu\orcidA{},~\IEEEmembership{Member,~IEEE},
Chengfang Li\orcidF{}, 
Rui Xi\orcidE{},
Wenhao Li\orcidB{},
Liang Cao\orcidH{},
Jin Wang\orcidD{},
Laurence T. Yang\orcidI{},~\IEEEmembership{Fellow,~IEEE},
Junsong Yuan\orcidJ{},~\IEEEmembership{Fellow,~IEEE},
Wei Zhou\orcidK{},~\IEEEmembership{Senior Member,~IEEE}
\IEEEcompsocitemizethanks{
\IEEEcompsocthanksitem This work is supported in part by the Mitacs Elevate Canada under Grant No. IT44479. (Corresponding author: Jing Liu)
\IEEEcompsocthanksitem Yang Liu is with College of Electronic and Information Engineering, Tongji University, Shanghai 201804, China (email: yangliu25@tongji.edu.cn)
\IEEEcompsocthanksitem Jing Liu is with the College of Future Information Technology, Fudan University, Shanghai 200433, China, also with the Division of Natural and Applied Sciences, Duke Kunshan University, Suzhou 215316, China, and also with the Department of Electrical and Computer Engineering, The University of British Columbia, BC V6T 1Z4, Canada (e-mail: jing.liu@ieee.org).
\IEEEcompsocthanksitem Chengfang Li is with the the Suzhou Institute of Biomedical Engineering and Technology (SIBET), Chinese Academy of Sciences, Suzhou 215163, China (e-mail: licf@sibet.ac.cn).
\IEEEcompsocthanksitem Rui Xi is with the Department of Electrical and Computer Engineering, The University of British Columbia, BC V6T 1Z4, Canada (e-mail: ruix@ece.ubc.ca).
\IEEEcompsocthanksitem Wenhao Li is with the School of Computer Science, University of Sydney, NSW 2006, Australia (e-mail: li58843972@163.com).
\IEEEcompsocthanksitem Liang Cao is with the Department of Chemical and Biological Engineering, The University of British Columbia, BC V6T 1Z4, Canada (e-mail: liangcao@mit.edu).
\IEEEcompsocthanksitem Jin Wang is with the School of Future Science and Engineering, Soochow University, Suzhou 215006, China (e-mail: wjin1985@suda.edu.cn).
\IEEEcompsocthanksitem Laurence T. Yang is with School of Computer Science and Artificial Intelligence, Zhengzhou University, Zhengzhou 450001, China, and also with the Department of Computer Science, St. Francis Xavier University, Antigonish, NS B2G 2W5, Canada (e-mail: ltyang@ieee.org).
\IEEEcompsocthanksitem Junsong Yuan is with Department of Computer Science and Engineering, State University of New York at Buffalo, Buffalo, NY 14260, USA (e-mail: jsyuan@buffalo.edu).
\IEEEcompsocthanksitem Wei Zhou is with the School of Computer Science and Informatics, Cardiff University, CF24 4AG Cardiff, U.K. (e-mail: zhouw26@cardiff.ac.uk).

}
}

\markboth{IEEE TRANSACTIONS ON PATTERN ANALYSIS AND MACHINE INTELLIGENCE}%
{Liu \MakeLowercase{\textit{\etal}}: Anomaly Detection and Generation with Diffusion Models: A Survey}

\IEEEtitleabstractindextext{

\begin{abstract}
    \justifying
  Anomaly detection (AD) plays a pivotal role across diverse domains, including cybersecurity, finance, healthcare, and industrial manufacturing, by identifying unexpected patterns that deviate from established norms in real-world data. Recent advancements in deep learning, specifically diffusion models (DMs), have sparked significant interest due to their ability to learn complex data distributions and generate high-fidelity samples, offering a robust framework for unsupervised AD. In this survey, we comprehensively review anomaly detection and generation with diffusion models (ADGDM), presenting a tutorial-style analysis of the theoretical foundations and practical implementations and spanning images, videos, time series, tabular, and multimodal data. Crucially, unlike existing surveys that often treat anomaly detection and generation as separate problems, we highlight their inherent synergistic relationship. We reveal how DMs enable a reinforcing cycle where generation techniques directly address the fundamental challenge of anomaly data scarcity, while detection methods provide critical feedback to improve generation fidelity and relevance, advancing both capabilities beyond their individual potential. A detailed taxonomy categorizes ADGDM methods based on anomaly scoring mechanisms, conditioning strategies, and architectural designs, analyzing their strengths and limitations. We final discuss key challenges including scalability and computational efficiency, and outline promising future directions such as efficient architectures, conditioning strategies, and integration with foundation models (e.g., visual-language models and large language models). By synthesizing recent advances and outlining open research questions, this survey (\href{https://github.com/fudanyliu/ADGDM}{\texttt{\faGithub ADGDM}}) aims to guide researchers and practitioners in leveraging DMs for innovative AD solutions across diverse applications.

\end{abstract}
}

\maketitle
\IEEEdisplaynontitleabstractindextext
\IEEEpeerreviewmaketitle

\section{Introduction}\label{sec1}

\IEEEPARstart{A}{nomaly} detection (AD) is a crucial task in various domains, from cybersecurity \cite{qiu2025selfsupervised} and finance \cite{livernoche2023diffusion} to healthcare \cite{pinaya2022fast} and industrial manufacturing \cite{yang2024novel}, aiming to identify instances deviating significantly from established normal patterns \cite{ho2025graph}.  However, the unsupervised nature of AD presents substantial challenges, particularly the lack of labeled anomalous data, which makes defining precise boundaries between normal and abnormal behavior difficult \cite{jin2024survey}. Traditional AD methods, often relying on predefined thresholds or assuming specific data distributions, are sensitive to noise and struggle to adapt to complex real-world scenarios \cite{ramachandra2022survey}. As data in modern applications become more complex and high-dimensional, more robust and adaptable techniques are needed \cite{zamanzadehdarban2025deep}. Consequently, innovative approaches are needed to learn from the inherent structure of normal data and effectively identify deviations without explicit supervision \cite{hojjati2024selfsupervised}.  The scarcity of labeled anomalies coupled with data heterogeneity complicates the development of effective unsupervised AD methods \cite{chang2023dataefficient}.  In addition, the need for interpretability in anomaly detection adds another layer of complexity, as understanding the reasons for flagged anomalies is crucial for effective decision-making \cite{carvalho2023invariant}.

Recent years have witnessed the emergence of DMs as a powerful class of generative models that have rapidly gained prominence across various computer vision tasks \cite{liu2025diffact,awais2025foundation}. At their core, DMs operate through a distinctive two-stage process: forward diffusion and reverse denoising. During forward diffusion, Gaussian noise progressively corrupts input data across multiple steps, transforming complex data distributions into simple isotropic Gaussian structures \cite{santos2023blackout}. The neural network-parameterized reverse denoising process then learns to iteratively remove the added noise, reconstructing original data from noisy samples while capturing intricate data dependencies \cite{duan2024soil}. Compared to traditional generative approaches such as generative adversarial networks (GANs) and variational autoencoders (VAEs), DMs offer significant advantages in training stability by avoiding mode collapse and vanishing gradient challenges \cite{kumar2023selfsupervised}. With superior sample quality and improved mode coverage, DMs generate diverse, high-fidelity samples that accurately represent underlying data distributions, positioning them at the forefront of generative modeling research for applications including anomaly detection \cite{li2023fast}.
Ongoing research addresses computational limitations through techniques like patching, which employs vision Transformer (ViT)-style transformations to reduce sampling time and memory requirements, while flexibility in the forward process enables multi-scale training that enhances overall performance \cite{qiu2025selfsupervised}.

Considering the inherent challenges of anomaly scarcity and data distribution complexity, identifying instances that deviate significantly from expected behavior poses particular challenges in unsupervised settings with limited labeled anomalies \cite{wang2023odd,wu2024vadclip}, where DMs provide a promising solution by learning complex data distributions and generating high-fidelity samples for effective reconstruction-based anomaly scoring~\cite{wang2024sparsedm,zhang2024safeguarding}. Specifically, a DM trained on normal data learns to progressively add and then reverse noise, effectively reconstructing data from a noise distribution \cite{tebbe2023d3ad}. Consequently, when presented with an anomalous input, the DM attempts to reconstruct a "normal" version, and the discrepancy between the input and its reconstruction, quantified as the reconstruction error, serves as an anomaly score \cite{yao2024glad,marimont2024disyre}.  A larger reconstruction error thus implies a higher likelihood of an anomaly, allowing DMs to capture subtle deviations, particularly in high-dimensional image data and industrial time-series monitoring \cite{tebbe2024dynamic,pintilie2023time,abanda2024reconstructionbased}.
A compact publication-and-citation trend summary is provided in the supplementary material in \cref{supp:fig:supp-trend-overview}, where it still shows the same central pattern: research interest in anomaly detection and diffusion models rises steeply from 2021 to 2025, with TSAD accounting for 41\% of the surveyed AD papers, followed by VAD (29\%) and IAD (22\%), indicating both broad modality coverage and especially strong momentum in sequential-data settings.
Cross-modality qualitative snapshots are also provided in the supplementary material \cref{supp:sec:supp-qualitative-examples} so that representative diffusion-based progress can be compared for image, video, time series, tabular, and multimodal AD.
Taken together, the distribution reflects both the versatility of diffusion models across different data modalities and the particularly challenging nature of detecting anomalies in sequential data~\cite{zhou2023dual,sui2024anomaly,yang2024balanced}. Due to incomplete data for the current year, the overall trend indicates sustained interest in applying DMs to anomaly detection and generation tasks across various domains.

\begin{table*}[t!]
  \centering
  \caption{Summary of our survey with recently AD surveys, highlighting the main focus, domain, scope, resources.}
  \label{tab:1}
  \vspace{-0.1cm}
  \setlength{\tabcolsep}{1mm}{
  \resizebox{0.99\textwidth}{!}{
\begin{tabular}{@{}cccccccccccccccc@{}}
    \toprule
 &  &  &  & \multicolumn{2}{c}{\textbf{Domain}} & \multicolumn{7}{c}{\textbf{Scopre}} & \multicolumn{3}{c}{\textbf{Resources}} \\ \cmidrule(l){5-6} \cmidrule(l){7-13} \cmidrule(l){14-16}
\multirow{-2}{*}{\textbf{Ref.}} & \multirow{-2}{*}{\textbf{Year}} & \multirow{-2}{*}{\textbf{Venue}} & \multirow{-2}{*}{\textbf{Main Foucs}} & Specific & General & IAD & VAD & TSAD & TAD & MAD & AG & DM & Dataset & Metric & Page \\ \midrule

    \cite{cook2020anomaly} & 2020 & IEEE IoTJ & {{Time series AD in IoT}} & $\checkmark$ &  & $\astrosun$ & $\astrosun$ & $\CIRCLE$ & $\Circle$ & $\Circle$ & $\astrosun$ & $\Circle$ & $\Circle$ & $\LEFTcircle$ & - \\
\cite{pang2021deepa} & 2021 & ACM CSUR & General AD with DL &  & $\checkmark$ & $\CIRCLE$ & $\LEFTcircle$ & $\CIRCLE$ & $\LEFTcircle$ & $\LEFTcircle$ & $\Circle$ & $\Circle$ & $\LEFTcircle$ & $\LEFTcircle$ & - \\
\cite{fernando2022deep} & 2022 & ACM CSUR & Medical AD & $\checkmark$ &  & $\CIRCLE$ & $\Circle$ & $\LEFTcircle$ & $\LEFTcircle$ & $\Circle$ & $\Circle$ & $\Circle$ & $\Circle$ & $\CIRCLE$ & - \\
\cite{ramachandra2022survey} & 2022 & IEEE TPAMI & Single-scene video AD & $\checkmark$ &  & $\Circle$ & $\CIRCLE$ & $\astrosun$ & $\astrosun$ & $\Circle$ & $\Circle$ & $\Circle$ & $\CIRCLE$ & $\CIRCLE$ & - \\
\cite{chandrakala2023anomaly} & 2023 & AI Rev. & DL for surveillance VAD & $\checkmark$ &  & $\Circle$ & $\CIRCLE$ & $\Circle$ & $\Circle$ & $\Circle$ & $\Circle$ & $\Circle$ & $\CIRCLE$ & $\LEFTcircle$ & - \\
\cite{ma2023comprehensive} & 2023 & IEEE TKDE & Graph AD with DL &  & $\checkmark$ & $\Circle$ & $\Circle$ & $\Circle$ & $\LEFTcircle$ & $\LEFTcircle$ & $\Circle$ & $\LEFTcircle$ & $\CIRCLE$ & $\CIRCLE$ & \href{https://xiaoxiaoma-mq. github.io/Awesome-Deep-Graph-Anomaly-Detection/}{\textcolor{blue}{\faGithub}} \\
\cite{ulhassan2023anomaly} & 2023 & IEEE COMST & Blockchain AD & $\checkmark$ &  & $\astrosun$ & $\astrosun$ & $\LEFTcircle$ & $\LEFTcircle$ & $\Circle$ & $\Circle$ & $\Circle$ & $\LEFTcircle$ & $\CIRCLE$ & - \\
\cite{kheddar2025reinforcement} & 2025 & IEEE COMST & RL-based intrusion detection review & $\checkmark$ &  & $\astrosun$ & $\astrosun$ & $\LEFTcircle$ & $\LEFTcircle$ & $\Circle$ & $\Circle$ & $\Circle$ & $\LEFTcircle$ & $\CIRCLE$ & - \\
\cite{jin2024survey} & 2024 & IEEE TPAMI & GNN for time series analytics &  & $\checkmark$ & $\Circle$ & $\Circle$ & $\CIRCLE$ & $\LEFTcircle$ & $\Circle$ & $\Circle$ & $\Circle$ & $\CIRCLE$ & $\CIRCLE$ & \href{https://github.com/KimMeen/Awesome-GNN4TS}{\textcolor{blue}{\faGithub}} \\
\cite{liu2024generalized} & 2024 & ACM CSUR & Generalized VAD taxonomy &  & $\checkmark$ & $\LEFTcircle$ & $\CIRCLE$ & $\Circle$ & $\Circle$ & $\LEFTcircle$ & $\Circle$ & $\Circle$ & $\CIRCLE$ & $\CIRCLE$ & \href{https://github.com/fudanyliu/GVAED.git}{\textcolor{blue}{\faGithub}} \\
\cite{liu2025networking} & 2024 & ACM CSUR & Networking systems for VAD & $\checkmark$ &  & $\LEFTcircle$ & $\CIRCLE$ & $\Circle$ & $\Circle$ & $\LEFTcircle$ & $\Circle$ & $\LEFTcircle$ & $\CIRCLE$ & $\CIRCLE$ & \href{https://github.com/fdjingliu/NSVAD}{\textcolor{blue}{\faGithub}} \\
\cite{zamanzadehdarban2025deep} & 2025 & ACM CSUR & Time series AD with DL &  & $\checkmark$ & $\Circle$ & $\Circle$ & $\CIRCLE$ & $\LEFTcircle$ & $\Circle$ & $\LEFTcircle$ & $\Circle$ & $\CIRCLE$ & $\CIRCLE$ & \href{https://github.com/zamanzadeh/ts-anomaly-benchmark}{\textcolor{blue}{\faGithub}} \\
\textit{Ours} & 2025 & - & AD \& AG with DMs &  & $\checkmark$ & $\CIRCLE$ & $\CIRCLE$ & $\CIRCLE$ & $\CIRCLE$ & $\CIRCLE$ & $\CIRCLE$ & $\CIRCLE$ & $\CIRCLE$ & $\CIRCLE$ & \href{https://github.com/fudanyliu/ADGDM.git}{\textcolor{blue}{\faGithub}} \\ \bottomrule
  \end{tabular}}}

  \fontsize{7pt}{2pt}\selectfont {
  \begin{minipage}{0.98\textwidth}
  \vspace{1px}
  \textit{Notes:} Symbol definitions used throughout the comparison: $\Circle$: "Not Covered", $\LEFTcircle$: "Partially Covered", $\astrosun$: "Not applicable", and $\CIRCLE$: "Fully Covered". Abbreviations include image AD (IAD), video AD (VAD), time series AD (TSAD), tabular AD (TAD), multimodal AD (MAD), anomaly generation (AG), and diffusion models (DMs).  
  \end{minipage}
  }
  \vspace{-15px}
\end{table*}

\noindent\textbf{Related Surveys.} Despite the growing interest in anomaly detection research, existing surveys typically focus on specific modalities or methodological aspects rather than offering a unified perspective across different data types and generative approaches. For example, Cook et al.~\cite{cook2020anomaly} concentrate solely on IoT time-series data anomaly detection without addressing generative approaches, while Pang et al.~\cite{pang2021deepa} provide a general overview of deep learning for anomaly detection but omit specific coverage of diffusion models. In the medical domain, Fernando et al.~\cite{fernando2022deep} thoroughly examine anomaly detection techniques but predominantly focus on discriminative rather than generative approaches. Regarding VAD, Ramachandra et al.~\cite{ramachandra2022survey} and Chandrakala et al.~\cite{chandrakala2023anomaly} offer valuable insights into single-scene surveillance applications but lack exploration of generative approaches such as diffusion models. Furthermore, while Liu et al.~\cite{liu2024generalized} establish a comprehensive taxonomy for VAD and Liu et al.~\cite{liu2025networking} investigate networking systems for VAD, neither addresses the potential of DMs. Similarly, Ma et al.~\cite{ma2023comprehensive} extensively explore graph anomaly detection with deep learning but without investigating diffusion-based approaches. More recently, Jin et al.~\cite{jin2024survey} and Zamanzadeh Darban et al.~\cite{zamanzadehdarban2025deep} have advanced TSAD research, yet their coverage of DMs remains limited. We further add cybersecurity-specific context by explicitly covering intrusion-detection and communication-network surveys such as Kheddar et al.~\cite{kheddar2025reinforcement}, because cyberattacks in industrial control systems, network telemetry, and structured security logs are important TSAD/TAD application settings. A detailed comparison between our survey and existing works is presented in \cref{tab:1}, which demonstrates our unique contribution in comprehensively addressing both anomaly detection and generation across multiple modalities through the lens of diffusion models.

\noindent\textbf{Scope and Contributions.}\label{sec1.4}
In this survey\footnote{ \textit{Resources at Github:} \url{https://github.com/fudanyliu/ADGDM}}, we provide a comprehensive and up-to-date examination of anomaly detection and generation with diffusion models (ADGDM), spanning multiple data modalities including images, videos, time series, and tabular data. A distinctive aspect is our integrated view of detection and generation as complementary processes, wherein DMs enable controllable generation of synthetic anomalies to address data scarcity, while detection insights guide targeted generation. Consequently, a synergistic feedback loop emerges, transforming the fundamental limitation of anomaly data scarcity into an opportunity for continuous improvement. As DMs continue to demonstrate remarkable capabilities in generative AI tasks across various domains, their application to anomaly detection represents a rapidly evolving frontier with significant potential. The key contributions of our survey are summarized as follows:

\begin{itemize}
    \item To the best of our knowledge, this is the first comprehensive survey specifically dedicated to examining the intersection of diffusion models with anomaly detection and generation across diverse data modalities including images, videos, time series, and tabular data. Our integrative perspective reveals common principles and modality-specific challenges, fostering knowledge transfer between previously siloed research areas.
    \item We establish a taxonomy that categorizes ADGDM methods based on their anomaly scoring mechanisms (reconstruction-based, density-based, score-based) and conditioning strategies (unconditional, conditional). Accordingly, the structured framework provides researchers with a clear understanding of the methodological landscape and facilitates the positioning of new contributions.
    \item Unlike previous surveys focusing solely on detection, we address both anomaly detection and generation with diffusion models, where the processes form a synergistic cycle in which generation creates diverse synthetic anomalies to overcome data scarcity, while detection techniques inform targeted generation approaches. We highlight how generative capabilities enhance data augmentation, model robustness testing, and self-supervised learning.
    \item We identify key challenges in the field, such as computational efficiency and identity shortcut problem, and propose promising research directions including efficient architectures, novel conditioning strategies, hybrid approaches integrating traditional techniques, and LLMs with diffusion-based frameworks. In turn, the roadmap aims to accelerate progress in this rapidly evolving field.
\end{itemize}

\begin{figure}[t!]
  \centering
  \includegraphics[width=0.48\textwidth]{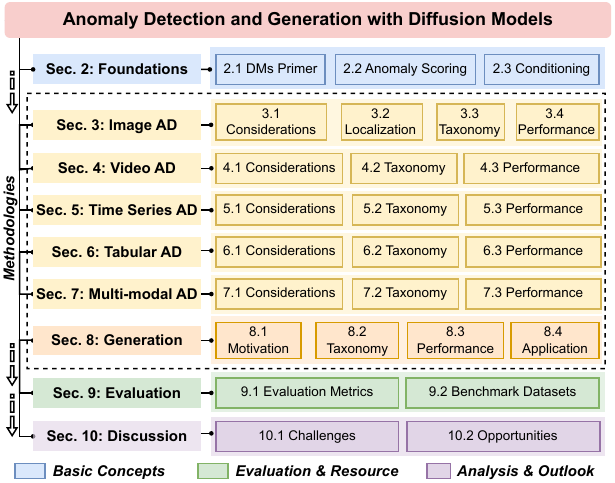}
  \caption{
Organization outline of this survey, covering the foundations of diffusion-based anomaly detection and generation, modality-specific anomaly detection, anomaly generation, evaluation, and future research opportunities.
  }
  \label{fig:2}
  \vspace{-15px}
\end{figure}

\noindent\textbf{Organization.}\label{sec1.5} 
As illustrated in \cref{fig:2}, the organization reflects the multifaceted nature of ADGDM research, highlighting both the core theoretical underpinnings and their practical applications across diverse domains. The remainder of this survey is structured as follows. \cref{sec2} establishes the foundations of ADGDM, covering diffusion models, anomaly scoring mechanisms, and conditioning strategies. \cref{sec3} and \cref{sec4} examine image and video AD respectively, addressing modality-specific challenges and methodological advances. \cref{sec5} and \cref{sec6} explore anomaly detection in time series and tabular data, respectively, while \cref{sec7} investigates multimodal anomaly detection. \cref{sec8} investigates anomaly generation with diffusion models. \cref{sec9} details evaluation metrics and benchmark datasets across different modalities. \cref{sec10} discusses open challenges and future research directions, and \cref{sec11} concludes this survey.

\section{Foundations}\label{sec2}
\subsection{DMs Primer}\label{sec2.1}

DMs constitute a powerful class of generative models capable of synthesizing high-quality data across various modalities \cite{mcallester2023mathematics}. At the core of DMs is a two-stage process: a forward diffusion process that gradually corrupts data by adding noise, and a reverse denoising process that learns to recover the original data from this noisy version. Mathematically, the forward process can be formulated as a stochastic differential equation (SDE) $d\mathbf{x} = f(\mathbf{x}, t)dt + g(t)d\mathbf{w}$ \cite{kidger2021neural} or its discrete-time counterpart, where Gaussian noise is iteratively added to an initial data sample $\mathbf{x}_0$ according to a predefined schedule $\beta_t$ \cite{cui2023elucidating}. In the discrete-time formulation, the forward process transforms $\mathbf{x}_0$ through a series of steps $q(\mathbf{x}_t|\mathbf{x}_{t-1}) = \mathcal{N}(\mathbf{x}_t; \sqrt{1-\beta_t}\mathbf{x}_{t-1}, \beta_t\mathbf{I})$, eventually approximating a standard Gaussian distribution $\mathcal{N}$. Sampling from DMs can be viewed as solving the corresponding SDEs or ordinary differential equations (ODEs) \cite{cui2023elucidating}, and recent work such as DynGMA \cite{zhu2024dyngma} explores robustly learning SDEs from data, potentially enhancing the forward process. In the reverse process, a neural network, typically a U-Net \cite{santos2023blackout}, is trained to predict either the noise $\boldsymbol{\epsilon}_\theta(\mathbf{x}_t, t)$ or estimate the score function $\nabla_{\mathbf{x}_t} \log p(\mathbf{x}_t)$ at each time step, effectively reversing the noise addition. During training, the network predicts either the original data or the added noise, guided by maximizing a variational lower bound on the data likelihood \cite{mcallester2023mathematics}. Several DM variants exist, including denoising diffusion probabilistic models (DDPMs) \cite{ghanem2024uncanny} which employ a Markov chain, denoising diffusion implicit models (DDIMs) \cite{cui2023elucidating} which offer a non-Markovian formulation for faster sampling, and SDE-based models \cite{kidger2021neural,dridi2021learning} for a more flexible framework. Additionally, alternative frameworks based on phase space dynamics \cite{chen2024generative} and methods using neural operators for accelerated sampling \cite{zheng2023fast} have been proposed. A mathematical analysis of singularities in DMs provides further insights into the challenges associated with learning complex data distributions, particularly when data lies on lower-dimensional manifolds \cite{lu2024mathematical}.

\subsection{Anomaly Scoring Mechanisms with DMs}\label{sec2.2}

Anomaly scoring with DMs leverages their learned understanding of data distributions to identify deviations from normality. DMs provide three primary paradigms for anomaly scoring, and the supplementary material retains the overview schematic so the main paper can focus on the scoring logic itself.

\subsubsection{Reconstruction-Based Scoring}\label{sec2.2.1}

Leveraging the reconstruction capability inherent in diffusion models, anomaly scoring can quantify the discrepancy between an input $\mathbf{x}$ and its reconstructed version $\hat{\mathbf{x}}$. Under this principle, a larger reconstruction error signifies a higher likelihood of $\mathbf{x}$ being anomalous, stemming from the model's training predominantly on normal data which hinders accurate reconstruction of out-of-distribution samples \cite{yao2025glad}. In the reconstruction paradigm, the process involves passing $\mathbf{x}$ through the reverse process to produce $\hat{\mathbf{x}}$, then comparing them. The reconstruction error can be formally expressed as $A(\mathbf{x}) = d(\mathbf{x}, \hat{\mathbf{x}})$, where common distance metrics $d$ include the $L_1$ norm, $L_2$ norm, or perceptual loss \cite{choi2022perception}. Perceptual loss uniquely compares high-level features extracted via pre-trained networks, capturing perceptual similarity rather than pixel-wise differences. Further refinements involve utilizing intermediate diffusion states $\mathbf{x}_t$ for scoring rather than just the final reconstruction. To refine scoring, intermediate diffusion states \(\mathbf{x}_t\) are utilized, capturing diverse data characteristics at varying noise levels \cite{hou2024highfidelity}. A generalized multi-step scoring approach aggregates errors over different timesteps as $\mathcal{S}(\mathbf{x}) = \sum{t=1}^{T} \mathbf{w}_t \mathcal{L}(\mathbf{x}, \hat{\mathbf{x}}_t)$, where weights $\mathbf{w}_t$ prioritize contributions from specific steps \cite{kumar2023selfsupervised}. Additionally, methods like dynamic denoising and latent space projections aim to improve reconstruction fidelity, particularly for anomalies of varying scales \cite{tebbe2024dynamic}.

\subsubsection{Density-Based Scoring}\label{sec2.2.2}

Density-based ADGDM exploits the model's capacity to learn the probability density function of the training data, operating under the premise that anomalies reside in low-density regions.  Specifically, an anomaly score is assigned based on the negative log-likelihood: $A(\mathbf{x}) = -\log p(\mathbf{x})$, where $A(\mathbf{x})$ represents the anomaly score and $p(\mathbf{x})$ is the estimated probability density of $\mathbf{x}$ \cite{xu2023unsupervised}.  A primary advantage of this approach is its interpretability, as the score directly corresponds to the probability density \cite{iwata2019supervised}.  However, accurately estimating this density, especially in high-dimensional spaces, poses a considerable challenge due to the curse of dimensionality, which necessitates a large number of samples for accurate estimation \cite{mikuni2024highdimensional}.  Furthermore, diffusion models often operate in a transformed space, where the estimated density might not accurately reflect the original data space density. Simultaneously, a key challenge stems from the fact that diffusion models primarily learn the score function (the gradient of the log density) rather than the density itself \cite{saremi2018deep}.  Estimating the density from the score function can be computationally intensive and error-prone, particularly in high dimensions.  Several studies have explored methods to address these challenges, including joint dimensionality reduction and density estimation using low-rank tensor models \cite{amiridi2022lowrank}, analyzing score approximation and distribution recovery in low-dimensional data \cite{chen2023score}, estimating the data manifold dimension \cite{stanczuk2023your}, and improving density estimation through maximum likelihood training \cite{song2021maximum}.
\subsubsection{Score-Based Scoring}\label{sec2.2.3}
Score-based anomaly detection utilizes the learned score function, $\nabla_{\mathbf{x}} \log p(\mathbf{x})$, representing the gradient of the log-probability density. $\nabla_{\mathbf{x}} \log p(\mathbf{x})$ provides crucial information about the local geometry of the probability density in data space \cite{marimont2024disyre}. Importantly, a higher score magnitude indicates steeper probability density change rather than directly indicating the probability value itself. The anomaly score is defined as $A(\mathbf{x}) = ||\nabla_{\mathbf{x}} \log p(\mathbf{x})||_2$.
The effectiveness stems from geometric properties of normal data manifolds, which typically form regions with smooth probability transitions, while anomalies often lie where density functions exhibit sharp changes. In practice, the pattern occurs when: (1) data points fall between normal clusters, causing gradients to point strongly toward nearby modes, or (2) out-of-distribution samples prompt stronger "pulling" toward the learned manifold.
During training, the model estimates scores at different noise levels \cite{han2024neural} through score matching. At each reverse diffusion step, the score function $\mathbf{s}_\theta(\mathbf{x}_t, t)$ guides denoising as $\mathbf{x}_{t-1} = \mathbf{x}_t - \epsilon \mathbf{s}_\theta(\mathbf{x}_t, t) + \sqrt{2\epsilon}\mathbf{z}$, pulling samples toward higher density regions \cite{permenter2023interpreting}. In AD, this represents restoring potentially anomalous inputs toward normality \cite{tebbe2023d3ad}, where persistent high score magnitudes signal anomalous behavior. Empirical studies confirm that while score magnitude and probability density aren't strictly monotonically related, score-based methods effectively identify anomalies through gradient field behavior near anomalous regions.

\subsection{Conditioning Strategies}\label{sec2.3}

Conditioning strategies are crucial for adapting diffusion models to anomaly detection by shaping the learned distribution and influencing reconstruction behavior.  We first examine unconditional diffusion models (UDMs), noting their simplicity but acknowledging potential limitations with complex data. Subsequently, we explore conditional diffusion models (CDMs), investigating how incorporating additional information, such as class labels, image content, or textual descriptions, improves anomaly detection and enables controllable anomaly generation.

\subsubsection{Unconditional DMs}\label{sec2.3.1}

UDMs learn the underlying data distribution without external labels, simplifying training and making them readily applicable to anomaly detection \cite{pintilie2023time}. In the standard formulation, an unconditional DM maps a standard Gaussian distribution $\mathcal{N}(\mathbf{0}, \mathbf{I})$ to the data distribution $p(\mathbf{x})$ through a forward Markov chain that gradually adds scheduled Gaussian noise, $q(\mathbf{x}_t | \mathbf{x}_{t-1}) = \mathcal{N}(\mathbf{x}_t; \sqrt{1 - \beta_t} \mathbf{x}_{t-1}, \beta_t \mathbf{I})$. Consequently, the learned reverse chain estimates $p_\theta(\mathbf{x}_{t-1} | \mathbf{x}_t) = \mathcal{N}(\mathbf{x}_{t-1}; \mu_\theta(\mathbf{x}_t, t), \Sigma_\theta(\mathbf{x}_t, t))$, where neural networks parameterize the mean and covariance used to denoise samples from noisy states. For anomaly detection, reconstruction mapping directs a test input toward its expected normal counterpart, and the discrepancy between the input and reconstruction, often measured by $L_1$ or $L_2$ norms, serves as the anomaly score \cite{tebbe2024dynamic}. Accordingly, a high reconstruction error indicates that the sample cannot be well explained by the learned normal distribution. However, the same simplicity also limits UDMs on heterogeneous or multimodal data because a single global normal distribution may underrepresent sub-population structure, reduce sensitivity to subtle deviations \cite{han2024diffusion}, and create identity shortcuts in which anomalous inputs are reconstructed too faithfully \cite{tebbe2023d3ad}, making such drawbacks especially problematic when normal data contain multiple operating modes, anatomical subgroups, or scene-specific variations. Consequently, UDMs remain valuable for relatively homogeneous normal data and transparent baseline construction, but complex settings typically require structured conditioning, stronger priors, adaptive noise schedules, or more localized scoring \cite{zuo2024unsupervised}.

\subsubsection{Conditional DMs}\label{sec2.3.2}

CDMs incorporate external information, denoted as $y$, to guide the denoising and thus control generated samples. In the reverse process, the neural network parameterizing the denoising process takes both the noisy input $\mathbf{x}_t$ and the conditioning signal $y$ as inputs, specifically $p_\theta(\mathbf{x}_{t-1}|\mathbf{x}_t, y) = \mathcal{N}(\mathbf{x}_{t-1}; \mu_\theta(\mathbf{x}_t, t, y), \Sigma_\theta(\mathbf{x}_t, t, y))$, where $\mu_\theta$ and $\Sigma_\theta$ represent the learned mean and covariance, respectively.  The conditioning signal $y$ can represent various data types depending on the application. For example, Zhan et al. \cite{zhan2024enhancing} use class labels as the conditioning signal for multi-class anomaly detection, allowing the model to learn class-specific distributions. Similarly, for image restoration, Mousakhan et al. \cite{mousakhan2023anomaly} condition the denoising process on the input image itself to reconstruct a defect-free version.  In another approach, Tebbe and Tayyub \cite{tebbe2024dynamic} utilize an initial anomaly prediction as a conditioning signal to guide the dynamic addition of noise during the forward diffusion process.  Additionally, Singh et al. \cite{singh2022conditioning} explore conditioning on crafted input noise artifacts for controlled image generation with semantic attributes.  For controllable anomaly generation, text-conditioned diffusion models, as explored by Sadat et al. \cite{sadat2024cads}, leverage textual descriptions to improve the diversity of generated samples. In the context of time series analysis, Pintilie et al. \cite{pintilie2023time} investigate the use of diffusion models for anomaly detection, while Tebbe and Tayyub \cite{tebbe2023d3ad} explore conditioning strategies for improved anomaly localization.

\section{Image Anomaly Detection}\label{sec3}

\subsection{Considerations}\label{sec3.1}

\begin{figure}[t!]
  \centering
\includegraphics[width=0.48\textwidth]{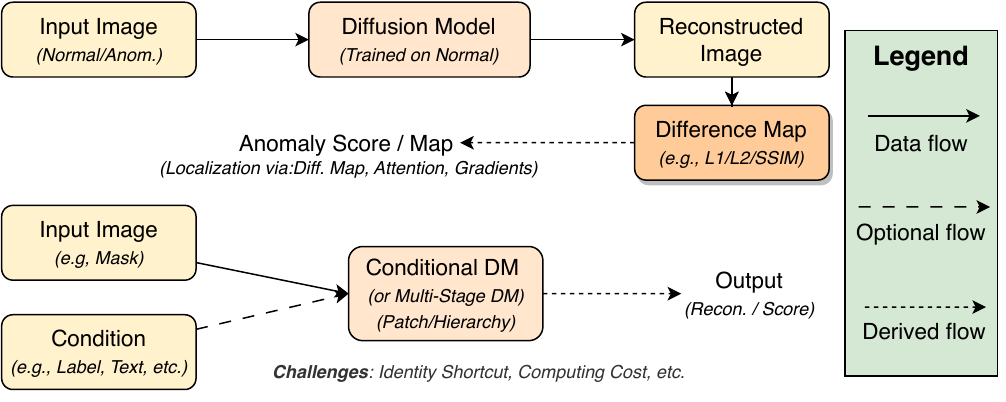}
  \caption{IAD with (a) reconstruction-based methods and (b) conditional/multi-stage variants.}
  \label{fig:4}
  \vspace{-15px}
\end{figure}

For IAD, the central design tension is between anomaly-sensitive reconstruction fidelity and practical denoising efficiency. As illustrated in \cref{fig:4}, identity shortcuts arise when diffusion models trained for high-fidelity reconstruction inadvertently reconstruct anomalous regions, suppressing the residual maps that should reveal defects \cite{yao2024glad,li2023fast}. Masked reconstruction mitigates this behavior by forcing the model to infer missing normal content \cite{wei2023diffusion}, feature editing steers anomalous latent representations toward normal counterparts \cite{hou2024highfidelity}, and adversarial training can improve robustness to perturbations \cite{lin2025adversarial}. Nevertheless, these remedies introduce different costs: masked reconstruction depends on reliable masking, feature editing requires meaningful latent-space control, and adversarial objectives add optimization complexity. Their usefulness also depends on whether the anomaly is localized, texture-like, structural, or distributed across an image. Computational demand forms a second practical constraint because iterative reverse diffusion can require 50--1000 denoising steps for high-resolution inspection, resulting in substantial latency for medical or industrial images. Acceleration strategies therefore shorten sampling or reduce representation complexity. Progressive distillation reduces the number of reverse steps while preserving reconstruction quality \cite{salimans2021progressive}. SLAM and FDM use linear ODE approximations or momentum-based acceleration \cite{xu2025accelerating,wu2023fast}, DPM-Solver streamlines the reverse process through efficient ODE solvers \cite{lu2022dpmsolver}, and LDMs reduce pixel-space cost by operating in lower-dimensional latent spaces \cite{liu2025dpldms}. LoRA and sparse-layer designs further improve deployability by limiting the number of trainable or active parameters \cite{golnari2023loraenhanced,wang2024sparsedm}. The resulting IAD design must therefore balance reconstruction sensitivity, localization quality, sampling time, and the availability of masks or other auxiliary priors.

\subsection{Anomaly Localization with DMs}\label{sec3.2}
Precise localization of anomalous regions is crucial for applications like industrial defect detection \cite{kumar2023selfsupervised} and medical diagnosis \cite{li2023fast}.  While diffusion models excel at capturing global data distributions, pinpointing anomalies requires specialized techniques, such as generating a difference map between the original and reconstructed image \cite{wyatt2022anoddpm}.  The magnitude of this difference then serves as the anomaly likelihood measure.  In addition, incorporating attention mechanisms within the diffusion model architecture offers finer-grained localization \cite{pinaya2022fast}, where attention weights highlight anomalous regions during reconstruction. Gradient-based methods provide another strategy \cite{che2024anofpdm}, identifying anomalous pixels by computing the gradient of the reconstruction loss with respect to the input.  Specifically, methods like AnoFPDM \cite{che2024anofpdm} leverage the forward diffusion process for brain MRI anomaly segmentation, while D3AD \cite{tebbe2023d3ad} uses dynamic denoising and latent projections to localize multi-scale anomalies, and AnoDDPM \cite{wyatt2022anoddpm} employs multi-scale simplex noise diffusion for improved localization accuracy, particularly in medical images.

\subsection{Taxonomy and Advances}\label{sec3.3}
IAD methods using diffusion models can be organized by scoring mechanism, conditioning strategy, and reconstruction design. As summarized in \cref{tab:2}, reconstruction-, density-, and score-based methods produce different anomaly evidence, while unconditional, conditional, single-stage, and multi-stage designs expose trade-offs among fidelity, calibration, auxiliary-prior availability, and deployment cost \cite{hu2024anomalydiffusion,liang2024modality,ahamed2024igcondapet,baugh2025imageconditioned,marimont2024disyre,mikuni2024highdimensional}. Additional details on reconstruction/density/score scoring, conditioning, and staging are provided in the supplementary material, \cref{supp:sec:supp-iad-taxonomy}.

\begin{table}[t!]
  \centering
  \caption{Summary of IAD methods across various imaging domains with implementations.}
  \label{tab:2}
  \vspace{-0.1cm}
  \setlength{\tabcolsep}{0.6mm}{
  \resizebox{0.48\textwidth}{!}{
    \begin{tabular}{@{}lcccc@{}}
      \toprule
      \textbf{Method} & \textbf{Year} & \textbf{Venue} & \textbf{Imaging Domains} & \textbf{Code} \\ \midrule
      Latent DDPM \cite{pinaya2022fast} & 2022 & MICCAI & Brain CT and MRI & - \\
      LDM-OOD \cite{graham2023unsupervised} & 2023 & MICCAI & 3D medical data & \href{https://github.com/marksgraham/ddpm-ood}{\textcolor{blue}{\faLink}} \\
      FNDM \cite{li2023fast} & 2023 & MICCAI & Brain MR images & - \\
      Cold Diffusion \cite{bansal2023cold} & 2023 & NeurIPS & General image & - \\
      Mao et al. \cite{mao2023guided} & 2023 & ACM MM & Text-image pairs & - \\
      THOR \cite{bercea2024diffusion} & 2024 & MICCAI & Brain MRI, Wrist X-rays & \href{https://github.com/compai-lab/2024-miccai-bercea-thor.git}{\textcolor{blue}{\faLink}} \\
      DDPM-DDIM \cite{fontanella2024diffusion} & 2024 & IEEE TMI & Brain images & - \\
      mDDPM \cite{iqbal2024unsupervised} & 2024 & MLMI & Brain MRI & \href{https://mddpm.github.io/}{\textcolor{blue}{\faLink}} \\
      MMCCD \cite{liang2024modality} & 2024 & MICCAIW & Multimodal MRI & - \\
      DAG \cite{navalmarimont2024ensembled} & 2024 & MICCAI & Brain MRI & - \\
      IgCONDA-PET \cite{ahamed2024igcondapet} & 2024 & ArXiv & Medical imaging (PET) & \href{https://github.com/igcondapet/IgCONDA-PET.git}{\textcolor{blue}{\faLink}} \\
      ODEED \cite{bellier2024detecting} & 2024 & CVPRW & Remote sensing & - \\
      AnomalyDiffusion \cite{hu2024anomalydiffusion} & 2024 & AAAI & Industrial images & \href{https://github.com/sjtuplayer/anomalydiffusion}{\textcolor{blue}{\faLink}} \\
      DualAnoDiff \cite{jin2024dualanodiff} & 2024 & ArXiv & Industrial images & - \\
      Babaei et al. \cite{babaei2024pancreatic} & 2024 & ArXiv & Medical imaging & - \\
      UNIMO-G \cite{li2024unimog} & 2024 & ACL & Text-image pairs & - \\
      ICDM \cite{baugh2025imageconditioned} & 2025 & USMLMI & Medical imaging & \href{https://github.com/matt-baugh/img-cond-diffusion-model-ad}{\textcolor{blue}{\faLink}} \\ \bottomrule
  \end{tabular}}}
  \vspace{-15px}
\end{table}

\subsection{Performance Comparison}\label{sec3.4}

Diffusion-based approaches for IAD have demonstrated significant performance improvements on standard benchmarks in recent years. As evidenced in \cref{tab:2key}, methods from 2021 to 2025 show a clear upward trajectory in detection metrics across challenging datasets like MVTec AD \cite{lu2023removing} and VisA \cite{yao2024glad}. Early approaches such as DRAEM \cite{zavrtanik2021draem} established strong foundations through discriminatively trained reconstruction embeddings for surface anomaly detection, while PaDiM \cite{defard2021padim} advanced the field through patch distribution modeling techniques. Subsequently, PatchCore \cite{roth2022total} achieved competitive results by leveraging memory banks of nominal patch features for efficient inference. Moving beyond traditional approaches, AnoDDPM \cite{wyatt2022anoddpm} introduced partial diffusion with simplex noise to outperform Gaussian diffusion models, particularly in medical image applications. Recent innovations include AutoDDPM \cite{bercea2023mask}, which enhances robustness through masking and resampling techniques, and Lu et al.'s \cite{lu2023removing} anomaly removal approach that treats irregularities as noise for improved localization. Notably, GLAD \cite{yao2024glad} represents the current state-of-the-art through its global-local adaptive diffusion framework, achieving nearly perfect detection scores on the VisA \cite{zou2022spotdifference}. Throughout the evolution of these methods, evaluation metrics have remained consistent, focusing on AUROC, AUPR, and pixel-level metrics like Dice coefficient and IoU for comprehensive assessment. Comparative analysis across these approaches reveals inherent trade-offs between detection accuracy, localization precision, and computational demands, with certain methods excelling in image-level anomaly detection (e.g., PatchCore on MVTec AD \cite{bergmann2019mvtec}) while others demonstrate superior anomaly localization capabilities (e.g., AnoDDPM~\cite{wyatt2022anoddpm} on medical images).
These comparisons are therefore interpreted as benchmark-specific evidence rather than a single absolute ranking, because localization-heavy medical settings and surface-inspection settings report different metric mixes, annotation granularity, and operational constraints.

\begin{table}[t!]
  \centering
  \caption{Performance comparison of key IAD methods on VisA and BTAD datasets.}
  \label{tab:2key}
  \vspace{-0.1cm}
  \setlength{\tabcolsep}{1.1mm}{
  \resizebox{0.48\textwidth}{!}{
    \begin{tabular}{@{}lcccc@{}}
      \toprule
      \multirow{2}{*}{\textbf{Method}}               & \multirow{2}{*}{\textbf{Year}} & \multirow{2}{*}{\textbf{Venue}} & \multicolumn{2}{c}{\textbf{Dataset (Metric)}} \\ \cmidrule(l){4-5} 
                                                     &                                &                                 & VisA (I-AUROC)          & BTAD (PRO)          \\ \midrule
      DRAEM \cite{zavrtanik2021draem}                & 2021                           & ICCV                            & 88.7, 73.1               & -                   \\
      SPADE \cite{cohen2021subimage}                 & 2021                           & Arxiv                           & 82.1, 65.9               & -                   \\
      PaDiM \cite{defard2021padim}                   & 2021                           & ICPR                            & 89.1, 85.9               & -                   \\
      RD4AD \cite{deng2022anomaly}                   & 2022                           & CVPR                            & 96.0, 70.9               & 94.3, 77.1           \\
      PatchCore \cite{roth2022total}                 & 2022                           & CVPR                            & 95.1, 91.2               & 92.7, 77.3           \\
      RAN \cite{lu2023removing}                      & 2023                           & ICCV                            & -                       & 93.4, 78.7           \\
      Tebbe et al. \cite{tebbe2024dynamic} & 2024                           & CVPRW                           & 96.0, 94.1               & 95.2, 83.2           \\
      GLAD \cite{yao2025glad}                        & 2025                           & ECCV                            & 99.5, 98.6               & -                   \\ \bottomrule
  \end{tabular}}}  
\vspace{-10px}
\end{table}

\section{Video Anomaly Detection}\label{sec4}

\subsection{Considerations}\label{sec4.1}
\begin{figure}[t!]
  \centering
\includegraphics[width=0.48\textwidth]{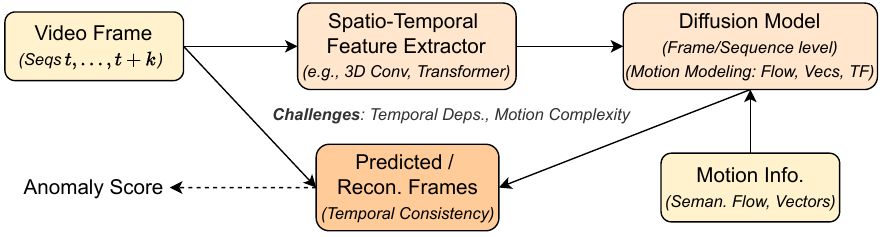}
  \caption{VAD incorporating spatio-temporal features and motion modeling.}
  \label{fig:5}
  \vspace{-10px}
\end{figure}

For VAD, informative deployment depends on modeling temporal context and motion strongly enough to distinguish true event-level anomalies from benign scene-specific variation \cite{lu2023vdt,park2024spectral}. As illustrated in \cref{fig:5}, VAD frameworks process sequential frames while capturing temporal dependencies, sudden trajectory changes, object interactions, and scene-level context \cite{zhang2019boosting,cong2013abnormal,jiang2011anomalous}. Diffusion models must therefore avoid treating videos as independent image collections when abnormality is expressed through action order, persistence, speed, or interactions rather than appearance alone. Spatio-temporal feature extractors and motion-conditioned diffusion modules help model these dependencies by incorporating flow vectors, past frames, or learned motion representations \cite{tur2023exploring,liu2025diffact}. Because an isolated action may be normal in one scene but anomalous in another, scene context and object interaction cues must be preserved together with motion dynamics. Recent approaches further combine spatio-temporal transformers such as VDT, motion-aware mechanisms such as SMA and MoVideo, and compressed-domain cues to preserve temporal consistency across clips \cite{awais2025foundation,liang2025movideo}. Robustness penalties and sampling-space truncation in VIDM can improve generation quality and anomaly discrimination \cite{mei2023vidm}, but these gains must be balanced against memory use, inference latency, scene-specific motion priors, and sensitivity to noisy optical-flow estimates. Consequently, practical VAD systems often trade fine-grained temporal modeling for scalable clip-level processing, especially in surveillance settings with long videos and sparse anomalous events.

\subsection{Taxonomy and Advances}\label{sec4.2}
As shown in \cref{tab:3}, VAD methods employ diffusion models across surveillance, traffic, and disaster-detection domains by varying processing granularity, motion modeling, and conditioning mechanisms. Frame-level designs scale efficiently but discard temporal evidence, whereas sequence-aware and motion-conditioned pipelines better capture evolving anomalies at the cost of higher memory, longer inference, and sensitivity to noisy optical-flow or scene-dependent motion priors \cite{li2026stnmamba,liu2025crcl,liu2025privacypreserving}. Additional details on frame-level processing, motion modeling, and conditioning strategies are provided in the supplementary material, \cref{supp:sec:supp-vad-taxonomy}.

\begin{table}[t!]
  \centering
  \caption{Summary of VAD with learning paradigms and available implementations.}
  \label{tab:3}
  \vspace{-0.1cm}
  \setlength{\tabcolsep}{0.45mm}{
  \resizebox{0.48\textwidth}{!}{
    \begin{tabular}{@{}lccccc@{}}
      \toprule
      \textbf{Method} & \textbf{Year} & \textbf{Venue} & \textbf{Type} & \textbf{Domain} & \textbf{Code} \\ \midrule
      Tur et al. \cite{tur2023exploring} & 2023 & IEEE ICIP & F & Video surveillance & \href{https://github.com/AnilOsmanTur/video_anomaly_diffusion}{\textcolor{blue}{\faLink}} \\
      Awasthi et al. \cite{awasthi2024anomaly} & 2024 & IEEE MMSP & F & Satellite imagery, disaster detection & - \\
      Video LDM \cite{blattmann2023align} & 2023 & CVPR & S, M & Driving simulation & - \\
      VidRD \cite{gu2023reuse} & 2023 & Arxiv & S & Text-to-video generation & - \\
      FPDM \cite{yan2023feature} & 2023 & ICCV & F & Video surveillance & - \\
      Stable Video Diffusion \cite{blattmann2023stable} & 2023 & Arxiv & S, M & Text/Image-to-video generation & - \\
      Wang et al. \cite{wang2023ensemble} & 2023 & Neurocomput. & F & Video surveillance & - \\
      Masked Diffusion \cite{fang2023masked} & 2023 & Arxiv & C & Video procedure planning & - \\
      GD-VDM \cite{lapid2023gdvdm} & 2023 & Arxiv & S & Complex scene video generation & - \\
      AADiff \cite{lee2023aadiff} & 2023 & Arxiv & C & Audio-aligned video synthesis & - \\
      Basak et al. \cite{basak2024diffusionbased} & 2024 & ESWA & F & Video surveillance & - \\
      DHVAD \cite{cheng2024denoising} & 2024 & IJCAI & F & Security and surveillance & - \\
      DiffVAD \cite{zhang2024safeguarding} & 2024 & IJCAI & F & Sustainable cities management & - \\
      GiCiSAD \cite{karami2024graphjigsaw} & 2024 & Arxiv & M & Skeleton-based VAD & - \\
      VADiffusion \cite{liu2024vadiffusion} & 2024 & Arxiv & F & Security surveillance & \href{https://github.com/LHaoooo/VADiffusion}{\textcolor{blue}{\faLink}} \\
      Cai et al. \cite{cai2024unsupervised} & 2024 & Sensors & F & Dust pollution monitoring & - \\
      FedDiff \cite{li2024feddiff} & 2024 & IEEE TCSVT & C & Multi-modal remote sensing & - \\ \bottomrule
  \end{tabular}}}

  \fontsize{6.5pt}{2pt}\selectfont {
  \begin{minipage}{0.48\textwidth}
  \vspace{1px}
  \textit{Notes: }"Type" refers to learning paradigm: F=Frame-level, S=Sequence-level, M=Motion Modeling, C=conditioning Strategies.
  \end{minipage}
  }
\vspace{-15px}
\end{table}

\subsection{Performance Comparison}\label{sec4.3}
Emerging research in VAD increasingly leverages diffusion models to capture complex temporal dynamics and generate high-fidelity video reconstructions. As shown in \cref{tab:3key}, contemporary methods demonstrate varying performance across standard benchmark datasets, with approaches like DMAD~\cite{liu2023diversitymeasurable} achieving exceptional results on UCSD Ped2 (99.7\% AUC) and CUHK Avenue (92.8\% AUC). Despite the promising capabilities of diffusion-based techniques, directly comparing model performance remains challenging due to inconsistent experimental protocols and reporting standards across studies. While traditional VAD research has predominantly employed autoencoders and generative adversarial networks, diffusion-based approaches like VADiffusion~\cite{liu2024vadiffusion} show competitive performance by employing a dual-branch architecture that combines motion vector reconstruction with I-frame prediction guided by compressed domain information. In turn, integrating diffusion principles enhances framework stability and detection accuracy, particularly when addressing both sudden and persistent anomalies. Nevertheless, computational overhead presents a significant concern for high-resolution video processing, creating an inevitable trade-off between detection accuracy and operational efficiency. Methods such as ReFLIP-VAD~\cite{dev2024reflipvad} and VadCLIP~\cite{wu2024vadclip}, though not directly via DMs, further highlight challenges in adapting complex vision-language architectures for video analysis despite their strong performance on datasets like XD-Violence \cite{wu2020not} and UCF-Crime  \cite{sultani2018realworld}.
For this reason, the comparison table is paired with surrounding analysis rather than an independent limitation column: the main weaknesses depend on whether a method targets frame reconstruction, temporal coherence, compressed-domain cues, or vision-language semantics.
\begin{table}[t!]
  \centering
  \caption{Performance (AUC) comparison of key VAD methods on UCSD Ped2, CUHK Avenue, and ShanghaiTech.}
  \label{tab:3key}
  \vspace{-0.1cm}
  \setlength{\tabcolsep}{0.5mm}{
  \resizebox{0.48\textwidth}{!}{
    \begin{tabular}{@{}lccccc@{}}
      \toprule
      \multirow{2}{*}{\textbf{Method}} & \multirow{2}{*}{\textbf{Year}} & \multirow{2}{*}{\textbf{Venue}} & \multicolumn{3}{c}{\textbf{Dataset}} \\ \cmidrule(l){4-6}
       &  &  & UCSD Ped2 & CUHK Avenue & ShanghaiTech \\ \midrule
      RTFM \cite{tian2021weaklysupervised} & 2021 & ICCV & 96.3 & 85.1 & 73 \\
      DMAD \cite{liu2023diversitymeasurable} & 2023 & CVPR & 99.7 & 92.8 & 78.8 \\
      LSH \cite{lu2023learnable} & 2023 & IEEE TCSVT & 91.3 & 87.4 & 77.6 \\
      VADNet \cite{huang2023boosting} & 2023 & IEEE TCSVT & - & 87.3 & 75.2 \\
      UMIL \cite{lv2023unbiased} & 2023 & CVPR & - & 88.3 & - \\
      UR-DMU \cite{zhou2023dual} & 2023 & AAAI & - & 88.9 & 97.92 \\
      VADiffusion \cite{liu2024vadiffusion} & 2024 & IEEE TCSVT & 98.2 & 87.2 & 71.7 \\
      PLOVAD \cite{xu2025plovad} & 2025 & IEEE TCSVT & - & - & 97.98 \\ \bottomrule
  \end{tabular}}}  
\vspace{-5px}
\end{table}

\section{Time Series Anomaly Detection}\label{sec5}
\begin{figure}[t!]
  \centering
\includegraphics[width=0.48\textwidth]{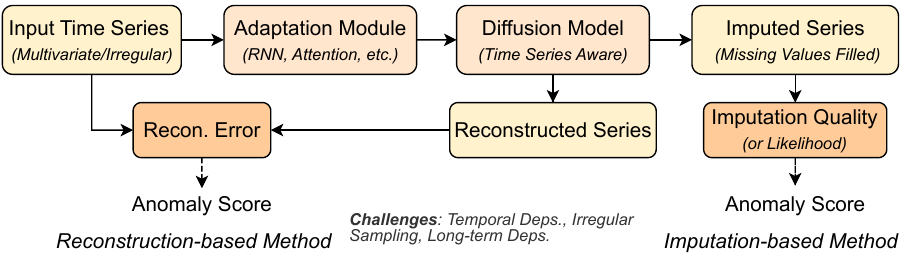}
  \caption{TSAD with reconstruction and imputation paths.}
  \label{fig:6}
  \vspace{-15px}
\end{figure}

TSAD with DMs faces distinct challenges related to the inherent temporal structure of sequential data. As shown in \cref{fig:6}, effective TSAD frameworks must address temporal dependencies, irregular sampling rates, and long-term relationships through specialized adaptation modules incorporating recurrent networks and attention mechanisms. The figure highlights two primary approaches in diffusion-based TSAD: reconstruction-based methods that compute anomaly scores by comparing original and reconstructed series, and imputation-based methods that evaluate anomaly likelihood through the quality of imputed missing values. Both approaches utilize specialized time series-aware diffusion models that capture the complex dynamics of sequential data, enabling more nuanced anomaly detection compared to traditional methods.
This setting is particularly important for cyberattack and intrusion detection, where multivariate telemetry from industrial control systems, cloud services, or network monitors must reveal stealthy deviations, false-data injection, and process manipulation without assuming abundant labeled attacks~\cite{kheddar2025reinforcement,goh2017dataset,su2019robust}.

\subsection{Considerations}\label{sec5.1}

For TSAD, the central issue is preserving long-range temporal evidence while keeping anomaly scores calibrated under irregular sampling, missing values, and online monitoring constraints \cite{sui2024anomaly,pintilie2023time}. Sequential anomalies may appear as short spikes, slow drifts, delayed correlations, or multi-sensor pattern changes, so diffusion models must combine generative reconstruction with temporal modules that capture relationships across extended horizons \cite{chen2024dynamic}. Recurrent networks such as LSTMs provide one route for modeling sequential dependencies \cite{ulhassan2023anomaly}, while attention mechanisms allow the model to focus on relevant windows during reconstruction or scoring \cite{pinaya2022fast}. Diffusion transformers such as TimeDiT extend this idea by coupling attention with generative sampling for high-quality temporal synthesis \cite{cao2024timedit}, and CAT uses a moment network with reinforcement learning to locate informative moments under irregular sampling \cite{hartvigsen2023finding}. Progressive learning can gradually increase data complexity and network capacity to represent long-term trends and short-term variations \cite{santos2023blackout}. Incomplete observations remain another practical obstacle, since real industrial or network telemetry often contains missing channels, unequal sampling intervals, and delayed measurements. Structural-prior-free diffusion designs and temporal self-attention mechanisms, including TranAD and Triformer, improve robustness and efficient long-sequence processing while preserving distinct temporal dynamics \cite{drakulic2022structured,tuli2022tranad,cirstea2022triformer}, which proves particularly critical for online monitoring, where a detector must react quickly without overfitting to transient noise or masking gradual process drift.

\subsection{Taxonomy and Advances}\label{sec5.2}
\begin{table}[t!]
  \centering
  \caption{Summary of TSAD with learning paradigms.}
  \label{tab:4}
  \vspace{-0.1cm}
  \setlength{\tabcolsep}{2mm}{
  \resizebox{0.48\textwidth}{!}{
    \begin{tabular}{@{}lccc@{}}
      \toprule
      \textbf{Method} & \textbf{Year} & \textbf{Venue} & \textbf{Learning Paradigm} \\ \midrule
      D$^3$R \cite{wang2023drift} & 2023 & NeurIPS & Decompo. and recon. \\
      DDMT \cite{yang2023ddmt} & 2023 & ArXiv & Mask-based \\
      DiffAD \cite{xiao2023imputationbased} & 2023 & KDD & Imputation-based \\
      SSSD \cite{alcaraz2023diffusionbased} & 2023 & ArXiv & Imputation and forecasting \\
      Diffusion+ \cite{yang2023diffusionbased} & 2023 & ESEC/FSE & Imputation-based \\
      SaSDim \cite{zhang2023sasdim} & 2023 & ArXiv & Noise-scaling diffusion \\
      Pintilie et al. \cite{pintilie2023time} & 2023 & ICDMW & Diffusion-based \\
      NetDiffus \cite{sivaroopan2024netdiffus} & 2024 & COMNET & Time-series imaging \\
      NGLS-Diff \cite{han2024diffusion} & 2024 & ECML PKDD & Latent space diffusion \\
      TimeADDM \cite{hu2024unsupervised} & 2024 & ICASSP & Diffusion-based \\
      Zuo et al. \cite{zuo2024unsupervised} & 2024 & APIN & Reconstruction-based \\
      TimeDiT \cite{cao2024timedit} & 2024 & ICMLW & Foundation model \\
      TSDE \cite{senane2024selfsupervised} & 2024 & ArXiv & Self-supervised learning \\
      ProDiffAD \cite{tian2024prodiffad} & 2024 & IJCNN & Progressive distillation \\
      TSAD-C \cite{ho2024contaminated} & 2024 & ArXiv & Graph conditional diffusion \\ \bottomrule
  \end{tabular}}}
\vspace{-15px}
\end{table}

As detailed in \cref{tab:4}, diffusion models for TSAD primarily follow reconstruction- and imputation-oriented architectures. Reconstruction-based methods specifically identify anomalies based on high reconstruction errors. DDTAD uses a 1-D U-Net within a DDPM framework for time-series reconstruction \cite{sui2024anomaly}, D$^3$R introduces dynamic decomposition to handle drifting series \cite{wang2023drift}, NGLS-Diff operates within a learned latent space of normal temporal patterns \cite{han2024diffusion}, and DDMT combines denoising diffusion, Transformer modules, and adaptive neighbor masking \cite{yang2023ddmt}, and while this approach remains intuitive because anomalous observations should be harder to reconstruct from a normal temporal model, excessive smoothing may weaken short or low-amplitude attack signatures. Imputation-based methods instead frame anomaly detection around missing-value recovery or density-ratio reasoning \cite{drakulic2022structured}. DiffAD uses density ratios and conditional weight-incremental diffusion for concentrated anomalous points \cite{xiao2023imputationbased}, SSSD combines conditional diffusion with structured state space models for imputation and forecasting \cite{alcaraz2023diffusionbased}, and Diffusion+ and SaSDim target cloud-failure prediction and spatial time-series imputation \cite{yang2023diffusionbased,zhang2023sasdim}. General-purpose diffusion transformers and graph conditional variants, including TimeDiT, TSDE, and TSAD-C, broaden the same paradigm to multiple temporal tasks \cite{cao2024timedit,senane2024selfsupervised,ho2024contaminated}. At the category level, reconstruction can smooth weak attack signatures \cite{yang2023diffusionbased}, imputation depends on missingness and calibration assumptions \cite{xiao2023imputationbased}, and graph/transformer variants improve context modeling while increasing cost for long horizons, dense telemetry, and latency-sensitive IDS pipelines \cite{zhou2023dual}.

\subsection{Performance Comparison}\label{sec5.3}
Originally developed for image generation, DMs are increasingly being adapted for TSAD, leveraging their capability to capture complex temporal dependencies and generate high-fidelity synthetic data \cite{su2019robust}. As evidenced in \cref{tab:4key}, methods demonstrate significant performance improvements across challenging benchmarks such as SWaT, WADI, MSL, and SMD datasets. Early approaches like DAGMM \cite{zong2018deep} established foundational techniques using deep generative models, while subsequent methods such as OmniAnomaly \cite{su2019robust} advanced the field through more sophisticated probabilistic modeling. Architectural adaptations addressing specific challenges of time series data have emerged, with researchers incorporating recurrent networks and attention mechanisms to handle irregular sampling and long-term dependencies \cite{wang2023drift}. Various conditioning strategies have been explored, including prototype-based methods \cite{li2023prototypeoriented} that achieve notable precision-recall balance on industrial datasets, and approaches conditioning on "nominality scores" \cite{lai2023nominality} that demonstrate improved performance in point-adjusted metrics. Applications span diverse domains from network monitoring \cite{xu2021anomaly} to industrial fault detection \cite{xu2023unsupervised}, with generative capabilities being leveraged for data augmentation and robustness enhancement \cite{tuli2022tranad}. Recent innovations such as D$^3$R \cite{wang2023drift} incorporate dynamic decomposition to handle unstable multivariate time series with data drift, while SensitiveHUE \cite{feng2024sensitivehue} represents the current state-of-the-art by enhancing sensitivity to normal patterns, achieving remarkable F1 scores (91.08\% on SWaT) and point-adjusted metrics (98.42\% on MSL). Comparative analysis across these approaches reveals an inherent trade-off between precision and recall, with certain methods excelling in different application domains based on their specific architectural characteristics and conditioning strategies.
Because TSAD papers often differ in point-adjustment policy, window length, contamination assumptions, and anomaly granularity, limitation analysis is summarized around protocol comparability and precision-recall trade-offs rather than forced into row-wise table cells.
\begin{table}[t!]
  \centering
  \caption{Performance comparison of key TSAD methods on SWaT, WADI, MSL, and SMD datasets.}
  \label{tab:4key}
  \vspace{-0.1cm}
  \setlength{\tabcolsep}{0.5mm}{
  \resizebox{0.48\textwidth}{!}{
    \begin{tabular}{@{}lcccccc@{}}      
      \toprule
      \multirow{2}{*}{\textbf{Method}} & \multirow{2}{*}{\textbf{Year}} & \multirow{2}{*}{\textbf{Venue}} & \multicolumn{4}{c}{\textbf{Dataset (Metrics)}} \\ \cmidrule(l){4-7} 
       &  &  & SWaT (P, R, F1, F1\_PA) & WADI (P, R, F1, F1\_PA) & MSL (P, R, F1, F1\_PA) & SMD (P, R, F1, F1\_PA) \\ \midrule
      DAGMM \cite{zong2018deep} & 2018 & ICLR & 27.46, 69.52, 39.37, 85.33 & 54.44, 26.99, 36.09, 61.65 & 25.91, 62.86, 36.69, 70.09 & 42.59, 50.46, 26.85, 72.29 \\
      OmniAnomaly \cite{su2019robust} & 2019 & ACM KDD & 98.25, 64.97, 78.22, 86.61 & 49.47, 12.98, 22.96, 41.72 & 16.19, 84.66, 27.18, 89.94 & 20.61, 46.73, 28.20, 75.29 \\
      AnomalyTran \cite{xu2021anomaly} & 2021 & ICLR & 12.00, 100.00, 21.43, 94.07 & 5.79, 43.43, 10.21, 89.10 &   -, -, 2.10, 93.59 & -, -,2.12, 92.33 \\
      TranAD \cite{tuli2022tranad} & 2022 & VLDB & 97.94, 60.52, 74.16, 91.04 & 86.88, 15.50, 26.00, 42.04 & 29.06, 75.96, 42.04, 94.94 & 26.95, 57.36, 37.16, 76.95 \\
      D$^3$R \cite{wang2023drift} & 2023 & NeurIPS & 12.04, 99.59, 21.49, 90.55 & 6.23, 18.93, 11.75, 35.46 & 11.04, 93.01, 19.74, 87.44 & 23.70, 52.63, 26.12, 95.09 \\
      PUAD \cite{li2023prototypeoriented} & 2023 & ICML & 97.94, 60.52, 74.16, 91.04 & 86.88, 15.50, 26.00, 42.04 & 29.06, 75.96, 42.04, 94.94 & 26.95, 57.36, 37.16, 76.95 \\
      NPSR \cite{lai2023nominality} & 2023 & NeurIPS & 93.42, 75.52, 83.52, 91.07 & 78.43, 50.33, 61.31, 75.22 & 24.03, 83.92, 37.37, 96.45 & 26.58, 62.36, 37.27, 76.95 \\
      SensitiveHUE \cite{feng2024sensitivehue} & 2024 & ACM KDD & 94.68, 87.74, 91.08, 96.75 & 86.51, 58.73, 69.96, 92.25 & 33.05, 71.26, 45.16, 98.42 & 29.54, 60.80, 39.76, 96.33 \\ \bottomrule
  \end{tabular}}}

    \fontsize{6.5pt}{2pt}\selectfont {
      \begin{minipage}{0.48\textwidth}
        \vspace{1px}
       \textit{Notes:} P, R, F1, and F1\_PA refer to Precision, Recall, F1-score, and Point-Adjusted F1-score.
      \end{minipage}
      }  
\vspace{-15px}
\end{table} 
\section{Tabular Anomaly Detection}\label{sec6}

\begin{figure}[t!]
  \centering
\includegraphics[width=0.48\textwidth]{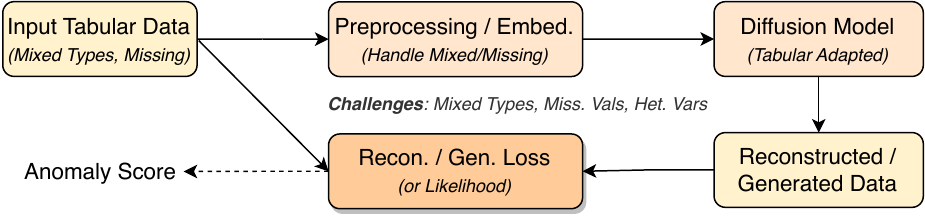}
  \caption{TAD handling mixed data types.}
  \label{fig:7}
  \vspace{-5px}
\end{figure}

Tabular data presents unique challenges for anomaly detection, primarily due to the inherent nature of mixed data types (numerical, categorical, ordinal) and the frequent presence of missing values. As illustrated in \cref{fig:7}, effective TAD frameworks must process heterogeneous input data through specialized preprocessing and embedding techniques before applying tabular-adapted diffusion models, which generate reconstructions that can be compared with the original input to compute anomaly scores. The figure highlights how reconstruction or generation loss serves as the primary mechanism for anomaly detection, while addressing key challenges specific to tabular data throughout the pipeline. Consequently, researchers have developed specific data preprocessing techniques and specialized diffusion model architectures to address these issues.
Beyond finance and healthcare, these design choices are also central to cybersecurity-oriented tabular AD, where flow records, alarm summaries, authentication logs, and other structured security events require anomaly scores that remain robust under heavy class imbalance, missing fields, and rapidly shifting attack distributions~\cite{kheddar2025reinforcement,han2022adbench}.

\subsection{Considerations}\label{sec6.1}
For TAD, informative evaluation depends on mixed-type modeling, missing-value handling, computational efficiency, and row-level provenance. However, since conventional diffusion models poorly handle categorical variables and heterogeneous feature scales, motivating DiSK for simultaneous processing of text, categorical, and continuous data \cite{kitouni2023disk}, while TabSynDex converts diverse feature types into unified latent representations \cite{chundawat2024tabsyndex}. Missing values further complicate tabular modeling. Consequently, methods such as MissDiff and DiffImpute address this limitation through masked denoising objectives or harmonized imputation \cite{ouyang2023missdiff,wen2024diffimpute}. Moreover, efficiency constraints have encouraged gradient-boosted score functions and accelerated sampling, with TimeAutoDiff illustrating parallel generation for time-series tabular data \cite{suh2024timeautodiff}. Additional adaptations use fair DMs to mitigate imbalance, Transformer denoisers to model feature interactions, latent diffusion to reduce representation complexity, and structured state space models for temporal tables \cite{yang2024balanced,zhang2023mixedtype,alcaraz2023diffusionbased}, driving practical deployment beyond finance and healthcare: fraud screening requires calibrated ranking under extreme imbalance \cite{livernoche2023diffusion}, clinical records require missing-value policies that preserve rare but actionable outliers \cite{kitouni2023disk}, and security-operation pipelines require provenance so analysts can trace which fields, joins, or retrieved exemplars triggered an alert \cite{kheddar2025reinforcement}. As a result, preprocessing choices, categorical encoders, missing-value policies, and post-hoc evidence extraction become as important as the diffusion backbone itself. A tabular detector with strong reconstruction accuracy may still be difficult to deploy if it cannot explain which fields drove an anomaly score or if its treatment of rare categories changes the operational meaning of an alert.

\subsection{Taxonomy and Advances}\label{sec6.2}
\begin{table}[t!]
  \centering
  \caption{Summary of TAD methods with type, metrics, and available implementations.}
  \label{tab:5}
  \vspace{-0.1cm}
  \setlength{\tabcolsep}{0.4mm}{
  \resizebox{0.48\textwidth}{!}{
    \begin{tabular}{@{}lccccc@{}}
      \toprule
      \textbf{Method} & \textbf{Year} & \textbf{Venue} & \textbf{Type} & \textbf{Performance Metrics} & \textbf{Code} \\ \midrule
      TabADM \cite{zamberg2023tabadm} & 2023 & ArXiv & Diffusion-based & Detection accuracy & - \\
      MPDR \cite{yoon2023energybased} & 2023 & NeurIPS & Energy-based & AUPR, AUROC & - \\
      Tritscher et al. \cite{tritscher2024generative} & 2024 & xAI & Generative inpainting & Explanation quality & - \\
      TimeAutoDiff \cite{suh2024timeautodiff} & 2024 & ArXiv & VAE + DDPM hybrid & Fidelity and utility metrics & \href{https://github.com/Hurongyao/TIMEADDM}{\textcolor{blue}{\faLink}} \\
      Dai et al. \cite{dai2025unsupervised} & 2025 & AAAI & Noise evaluation & AUC score & - \\
      CoDi \cite{lee2023codi} & 2023 & ICML & Co-evolving diffusion & Synthesis quality & \href{https://github.com/ChaejeongLee/CoDi}{\textcolor{blue}{\faLink}} \\
      SimpDM \cite{liu2024selfsupervision} & 2024 & CIKM & Self-supervised diffusion & Imputation accuracy & - \\
      FinDiff \cite{sattarov2023findiff} & 2023 & ICAIF & Diffusion-based & Fidelity, privacy, utility & - \\
      PHAD \cite{li2025prototypeoriented} & 2025 & IPM & Hypergraph representation & Detection accuracy & - \\
      Thimonier et al. \cite{thimonier2024retrieval} & 2024 & CIKM & Retrieval-augmented & Detection performance & - \\ \bottomrule
  \end{tabular}}}

  \fontsize{6.5pt}{2pt}\selectfont {
  \begin{minipage}{0.48\textwidth}
  \textit{Notes:} Public implementations are listed when directly available from the cited methods.
  \end{minipage}
  }
\vspace{-10px}
\end{table}

Modern approaches to TAD with generative models exhibit distinct methodological divisions, as illustrated in \cref{tab:5}, which categorizes recent techniques across multiple dimensions including architectural paradigms and evaluation frameworks. Among reconstruction-based methods, TabADM~\cite{zamberg2023tabadm} leverages diffusion principles to model normal data distributions and identify anomalies through reconstruction errors, establishing a foundational approach for diffusion-based tabular analysis. Complementary to this paradigm, generation-based techniques such as FinDiff~\cite{sattarov2023findiff} and CoDi~\cite{lee2023codi} focus on synthesizing realistic tabular data, enabling broader applications spanning data augmentation and comparative anomaly detection between generated and real instances. Beyond standard diffusion approaches, self-supervised paradigms exemplified by SimpDM~\cite{liu2024selfsupervision} enhance imputation accuracy and robustness when handling sparse or missing values, a prevalent challenge in tabular domains. Integration with complementary techniques further diversifies the methodological landscape, with PHAD~\cite{li2025prototypeoriented} merging diffusion-based augmentation with hypergraph representation learning to capture higher-order relationships, while retrieval-augmented methods~\cite{thimonier2024retrieval} leverage retrieved samples for enhanced anomaly identification. Architectural innovations continue to emerge through hybrid approaches like TimeAutoDiff~\cite{suh2024timeautodiff}, which synergistically combines autoencoder frameworks with DMs to enhance data synthesis by capitalizing on the strengths of both paradigms.
Across these categories, diffusion-based tabular models still depend heavily on the choice of categorical embeddings and missing-value handling~\cite{sattarov2023findiff}, retrieval or hypergraph augmentations increase memory/pipeline complexity~\cite{thimonier2024retrieval}, and generation-oriented methods can improve robustness but do not automatically yield interpretable row-level evidence for high-stakes decisions such as fraud or intrusion screening~\cite{suh2024timeautodiff}.
\begin{table}[t!]
  \centering
  \caption{Performance comparison of key TAD models on 47 real-world tabular datasets, including domains such as healthcare, image processing, and finance.}
  \label{tab:5key}
  \vspace{-0.1cm}
  \setlength{\tabcolsep}{0.5mm}{
  \resizebox{0.48\textwidth}{!}{
    \begin{tabular}{@{}lccccc@{}}
      \toprule
      
      \multirow{2}{*}{\textbf{Method}} & \multirow{2}{*}{\textbf{Year}} & \multirow{2}{*}{\textbf{Venue}} & \multicolumn{3}{r}{\textbf{Avg. performance on 47-dataset}} \\ \cmidrule(l){4-6} 
       &  &  & AUC (\%) & Mean Rank & p-value \\ \midrule
      NeuTraLAD \cite{qiu2021neural} & 2021 & ICML & 71.45 ± 22.6 & 9.16 & 0 \\
      SCAD \cite{shenkar2021anomaly} & 2022 & ICLR & 69.15 ± 15.3 & 6 & 0.0002 \\
      ECOD \cite{li2023ecod} & 2023 & IEEE TKDE & 52.77 ± 11.7 & 10.12 & 0 \\
      PLAD \cite{cai2022perturbation} & 2022 & NeurIPS & 67.42 ± 19.6 & 9.16 & 0 \\
      DPAD \cite{yuan2024dpad} & 2024 & Arxiv & 88.05 ± 12.4 & 4.96 & 0.0003 \\
      AutoUAD \cite{dai2024autouad} & 2024 & Arxiv & 92.68 ± 11.8 & 2.04 & 0.47 \\
      DNN \cite{dai2025unsupervised} & 2025 & AAAI & 92.27 ± 11.1 & 1.68 & 0.47 \\ \bottomrule
         
  \end{tabular}}}  
\vspace{-5px}
\end{table}

\subsection{Performance Comparison}\label{sec6.3}
Direct application of image-based DMs to tabular data presents challenges due to inherent structural differences, including mixed data types and missing values, which necessitate specialized adaptations and preprocessing techniques. As evidenced in \cref{tab:5key}, TAD approaches demonstrate considerable performance variations across 47 real-world datasets spanning healthcare, finance, and image processing domains. Earlier methods like NeuTraLAD \cite{qiu2021neural} and SCAD \cite{shenkar2021anomaly} established foundational techniques with AUC scores of 71.45\% and 69.15\% respectively, while recent innovations including DPAD \cite{yuan2024dpad} and AutoUAD \cite{dai2024autouad} show substantial improvements, achieving 88.05\% and 92.68\% AUC scores. Related work in multivariate time series demonstrates competitive performance through specialized architectures that capture complex dependencies in structured data \cite{pintilie2023time}. Advanced frameworks incorporating dynamic step size computation and latent space projection \cite{tebbe2024dynamic} enhance both detection accuracy and anomaly localization capabilities. Applications span diverse domains from fraud detection to intrusion monitoring, leveraging diffusion models' ability to learn complex distributions for identifying unusual patterns in tabular structures. Notably, few-shot frameworks like AnomalySD \cite{yan2024anomalysd} utilizing Stable Diffusion have demonstrated particular effectiveness in industrial settings, achieving high performance with minimal labeled examples. Comparative analysis across these approaches reveals the importance of domain-specific adaptations, with methods like DNN \cite{dai2025unsupervised} achieving state-of-the-art performance (92.27\% AUC) through specialized architectures tailored to tabular data characteristics.
The table should be read together with the taxonomy discussion: tabular methods vary in feature types, missing-value policy, and label imbalance, so their weaknesses are more reliably described as category-level constraints than as row-wise claims detached from the original experimental settings.

\section{Multimodal Anomaly Detection}\label{sec7}

\begin{figure}[t!]
  \centering
\includegraphics[width=0.48\textwidth]{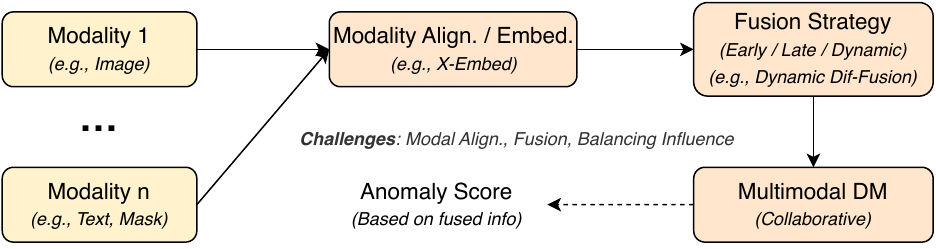}
  \caption{MAD with early/late/dynamic fusion strategies.}
  \label{fig:8}
  \vspace{-15px}
\end{figure}

By leveraging the complementary nature of diverse data sources, multimodal approaches significantly enhance the accuracy and robustness of anomaly detection systems. \cref{fig:8} illustrates how these frameworks process multiple input modalities through alignment and embedding techniques before applying fusion strategies (early, late, or dynamic) within collaborative diffusion models. The diagram demonstrates data flow while highlighting critical challenges including modal alignment, fusion techniques, and influence balancing.

\subsection{Considerations}\label{sec7.3.1}

For MAD, the most informative challenge axis is how alignment, fusion, and missing-modality handling affect whether cross-modal disagreement becomes usable anomaly evidence or just another source of false alarms~\cite{tebbe2024dynamic,liu2025multimodal}.
MAD introduces challenges due to diverse data sources. Modal alignment is challenging as different modalities occupy distinct feature spaces with varying scales and distributions \cite{costanzino2024multimodal}. Effective fusion techniques are crucial for capturing modality interactions, as simple concatenation may not exploit their complementary nature \cite{roheda2021robust}. Additionally, balancing the influence of modalities is essential, as some may be more informative for specific anomaly types \cite{zhang2023provable}. To address these challenges, collaborative diffusion frameworks enable joint learning of representations across modalities, promoting alignment and facilitating information exchange \cite{hu2024anomalyxfusion}.  Multimodal embedding spaces provide a common ground for representing and comparing information, mitigating the alignment issue, as exemplified by AnomalyXFusion's X-embedding approach, which integrates image, text, and mask features \cite{chen2025improving}.  Dynamic fusion mechanisms, such as AnomalyXFusion's Dynamic Dif-Fusion module, offer adaptive weighting of modalities based on their relevance, adjusting the contribution of multimodal features during diffusion steps for context-aware generation \cite{chen2023interpretation}.  Ongoing research on cross-modal feature mapping and quality-aware fusion further enhances the robustness and accuracy of these systems \cite{costanzino2024multimodal,zhang2023provable}.  Deep structured anomaly detection frameworks \cite{li2019deepa} and the incorporation of semantic guidance in diffusion models \cite{he2023diad} also contribute to handling the complexities of multimodal data.
Compared with image-only or video-only AD, multimodal deployment further requires deciding when disagreement between modalities is itself anomalous, when one modality should override another, and how to continue operating when text, audio, masks, or metadata are delayed or missing~\cite{chen2024counterfactual,hu2024anomalyxfusion}. As a result, calibration, fallback design, and human-in-the-loop validation become central parts of MAD pipelines rather than implementation details.

\subsection{Taxonomy and Advances}\label{sec7.3.2}

Recent research in anomaly detection increasingly exploits complementary information from diverse data sources to enhance detection accuracy and robustness. As shown in \cref{tab:6}, recent multimodal approaches leverage various fusion strategies tailored to specific application domains and data characteristics. Early fusion methods, exemplified by FedDiff \cite{li2024feddiff}, combine modalities at the input level, integrating features before model processing. In contrast, late fusion approaches like Ano-cDiff \cite{chen2024counterfactual} process each modality independently before combining outputs, preserving modality-specific characteristics. Dynamic fusion strategies, represented by Collaborative Diffusion \cite{huang2023collaborative}, adaptively weight modality contributions based on input characteristics, optimizing information extraction from heterogeneous sources. Multiple modality combinations have demonstrated effectiveness across application domains, with image-text pairings \cite{capogrosso2024exploiting} enhancing industrial defect detection through expert knowledge integration, video-audio combinations \cite{flaborea2023multimodal,lee2023aadiff} improving surveillance anomaly recognition, and image-mask pairings \cite{hu2024anomalyxfusion} facilitating precise region localization. Notable architectural innovations include X-embeddings \cite{hu2024anomalyxfusion} that unify features from multiple modalities into coherent embedding spaces, and human-in-the-loop systems \cite{capogrosso2024exploiting} that incorporate expert knowledge through text descriptions and region localization, substantially enhancing model performance on datasets like KSDD2. The growing availability of open-source implementations for methods like AnomalyXFusion and Ano-cDiff further accelerates progress in this rapidly evolving research area.
However, early fusion can blur modality-specific anomaly signatures, late fusion depends on well-calibrated per-modal scorers, and dynamic fusion remains vulnerable to misalignment, missing modalities, or one dominant signal overwhelming weaker but safety-critical evidence~\cite{li2019deepa,shankar2021neural}.

\begin{table}[t!]
\centering
\caption{Summary of MAD methods on different datasets with available implementations.}
\label{tab:6}
\setlength{\tabcolsep}{0.3mm}{
\resizebox{0.48\textwidth}{!}{
  \begin{tabular}{@{}lcccc@{}}
    \toprule
    {\textbf{Method}} & {\textbf{Year}} & {\textbf{Venue}} & {\textbf{Datasets}} & {\textbf{Code}} \\ \midrule
    Multimodal LDM \cite{capogrosso2024exploiting} & 2024 & Ital-IA & KSDD2 & - \\
    Ano-cDiff \cite{chen2024counterfactual} & 2024 & ESWA & Brain glioma datasets & \href{https://github.com/Snow1949/Ano-cDiff}{\textcolor{blue}{\faLink}} \\
    Flaborea et al. \cite{flaborea2023multimodal} & 2023 & ICCV & UBnormal, HR-UBnormal, etc., & - \\
    AnomalyXFusion \cite{hu2024anomalyxfusion} & 2024 & ArXiv & MVTec AD, LOCO, MVTec Caption & \href{http://github.com/hujiecpp/MVTec-Caption}{\textcolor{blue}{\faLink}} \\
    Collaborative Diffusion \cite{huang2023collaborative} & 2023 & CVPR & CelebAMask-HQ, CelebA-Dialog & - \\ \bottomrule
\end{tabular}}}
\vspace{-5px}
\end{table}

\subsection{Applications}\label{sec7.3.3}

MAD with DMs offers substantial advancements in Industry 5.0, especially for defect detection.  In this setting, diffusion models integrate diverse data modalities like images and text, providing a more comprehensive understanding of industrial processes and facilitating accurate anomaly identification.  Approaches combining expert knowledge with visual data show particular promise. For example, multimodal LDMs incorporate text descriptions and region localization \cite{capogrosso2024exploiting}, enabling more targeted anomaly detection. In these models, experts provide textual descriptions and localize potential anomalies within images, guiding the model towards learning relevant and interpretable defect representations \cite{capogrosso2024exploiting}.  Consequently, the integration of human expertise enhances the model's ability to detect subtle anomalies often missed by purely data-driven methods, and facilitates a robust feedback loop for iterative refinement and improved accuracy \cite{capogrosso2024exploiting}. Similarly, the Myriad model \cite{li2025myriad} uses an Expert Perception module to embed prior knowledge as tokens understandable by LLMs \cite{capogrosso2023neurosymbolic}. Adopting this neuro-symbolic strategy facilitates the generation of granular anomaly descriptions encompassing attributes like color, shape, and category \cite{li2025myriad}, which is crucial for effective decision-making in industrial settings.
Broader MAD case studies suggest the same pattern beyond industrial inspection: medical image-report pairs benefit from cross-checking lesion evidence against textual findings~\cite{capogrosso2024exploiting}, audio-visual surveillance relies on asynchronous cues for rare events~\cite{wei2022msaf}, and cyber-physical monitoring increasingly combines telemetry with logs or alerts so operators can verify whether a cross-modal inconsistency reflects a true incident or a benign sensor failure~\cite{goh2017dataset}.

\section{Anomaly Generation}\label{sec8}

\subsection{Motivation}\label{sec8.1}

\begin{figure}[t!]
  \centering
\includegraphics[width=0.48\textwidth]{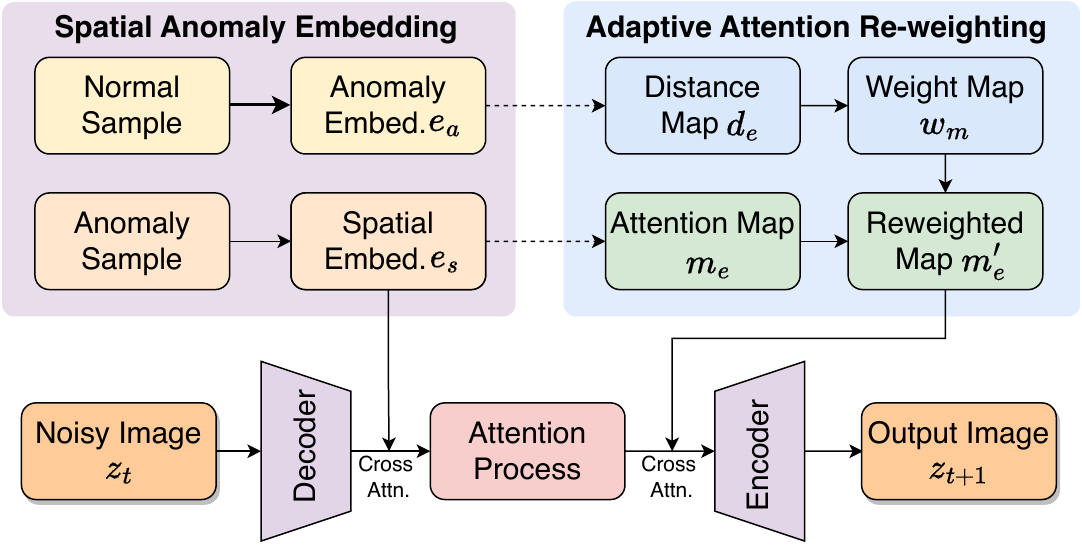}
\caption{Overview of the AnomalyDiffusion framework.}
  \label{fig:10}
  \vspace{-15px}
\end{figure}

\begin{figure}[t!]
  \centering
\includegraphics[width=0.48\textwidth]{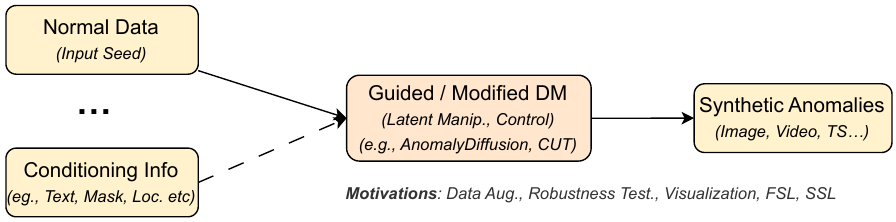}
  \caption{AG techniques with guided DMs for data augmentation, robustness testing, and self-supervised learning.}
  \label{fig:9}
  \vspace{-5px}
\end{figure}

Synthetic anomaly generation with DMs addresses the scarcity, imbalance, and limited diversity of real anomalies while enabling data augmentation, robustness testing, and self-supervised training. As shown in \cref{fig:9}, AG methods typically start from normal data seeds and optional conditions such as text descriptions or masks, then use guided diffusion to synthesize anomalies across modalities. Models such as CUT manipulate latent spaces to create diverse and realistic visual anomalies without requiring large anomaly collections \cite{sun2024cut}. As illustrated in \cref{fig:10}, AnomalyDiffusion further integrates spatial anomaly embeddings and adaptive attention re-weighting to control anomaly location and appearance more precisely \cite{hu2024anomalydiffusion}. Beyond expanding training sets, controllable generation supports stress testing by varying anomaly size, shape, severity, and context, which helps expose detector vulnerabilities in settings such as medical imaging \cite{tebbe2024dynamic,tebbe2023d3ad,celaya2024generalized}. Systematically generated anomalies can also reveal failure modes under rare or extreme conditions that are difficult to collect in real deployments, such as small defects, unusual lesion shapes, or low-contrast abnormal regions. Because the generation process can be conditioned, AG also enables controlled ablation studies that isolate whether a detector is sensitive to location, texture, scale, or semantic category. Finally, generated anomalies support self-supervised learning by mixing normal samples with synthetic abnormal patterns, allowing models to learn discriminative cues without dense manual labels \cite{kumar2023selfsupervised,li2024selfsupervised}.

\subsection{Taxonomy and Advances}\label{sec8.2}

Various diffusion-based approaches for anomaly generation have emerged in recent years, each addressing different aspects of the generation challenge with distinct methodological foundations. As summarized in \cref{tab:7}, these methods can be classified according to their underlying techniques and application domains. Manipulation-based approaches directly alter the diffusion process to produce anomalous samples, while conditional generation methods leverage additional information to guide the generation toward specific anomaly types. For example, AnomalyDiffusion \cite{hu2024anomalydiffusion} and DualAnoDiff \cite{jin2024dualanodiff} utilize specialized conditioning strategies for industrial inspection applications, whereas CUT \cite{sun2024cut} offers a more general framework for visual anomaly detection. Latent space modification represents another significant category, where techniques modify the underlying representation space to create anomalies, contrasting with data-space approaches that operate directly on raw input. Furthermore, the integration of diffusion models with other generative or discriminative architectures has gained traction, as exemplified by AdaBLDM \cite{li2024novel}, which enhances the realism and diversity of generated anomalies. Beyond visual domains, applications have expanded to specialized fields such as financial data analysis with FinDiff \cite{sattarov2023findiff} and network traffic analysis through NetDiffus \cite{sivaroopan2024netdiffus}, demonstrating the versatility of diffusion models for anomaly generation across diverse domains. The evolution from earlier techniques like CutPaste \cite{li2021cutpaste} to sophisticated diffusion-based frameworks indicates significant progression in generating increasingly realistic and controllable anomalies.

\begin{table}[t!]
  \centering
  \caption{Summary of AG methods across various domains with implementations.}
  \label{tab:7}
  \vspace{-0.1cm}
  \setlength{\tabcolsep}{0.4mm}{
  \resizebox{0.48\textwidth}{!}{
    \begin{tabular}{@{}lccccc@{}}
      \toprule
      {\textbf{Method}} & {\textbf{Year}} & {\textbf{Venue}} & {\textbf{Domain}} & \textbf{DM} & {\textbf{Code}} \\ \midrule
      HumanRefiner \cite{fang2024humanrefiner} & 2024 & ECCV & Human image generation & \ding{51} & \href{https://github.com/Enderfga/HumanRefiner}{\textcolor{blue}{\faLink}} \\
      AnomalyDiffusion \cite{hu2024anomalydiffusion} & 2024 & AAAI & Industrial inspection & \ding{51} & \href{https://github.com/sjtuplayer/anomalydiffusion}{\textcolor{blue}{\faLink}} \\
      DualAnoDiff \cite{jin2024dualanodiff} & 2024 & ArXiv & Industrial inspection & \ding{51} & \href{https://doi.org/10.48550/arXiv.2408.13509}{\textcolor{blue}{\faLink}} \\
      AdaBLDM \cite{li2024novel} & 2024 & ArXiv & Industrial defect generation & \ding{51} & \href{https://github.com/GrandpaXun242/AdaBLDM.git}{\textcolor{blue}{\faLink}} \\
      CUT \cite{sun2024cut} & 2024 & ArXiv & Visual anomaly detection & \ding{51} & \href{https://doi.org/10.48550/arXiv.2406.01078}{\textcolor{blue}{\faLink}} \\
      Rai et al. \cite{rai2024video} & 2024 & CVPR & Video anomaly detection & \ding{51} & - \\
      NSA \cite{schluter2022natural} & 2022 & ECCV & Manufacturing, Medical imaging & \ding{55} & \href{https://github.com/hmsch/natural-synthetic-anomalies}{\textcolor{blue}{\faLink}} \\
      CutPaste \cite{li2021cutpaste} & 2021 & CVPR & Industrial inspection & \ding{55} & - \\
      PRN \cite{zhang2023prototypical} & 2023 & CVPR & Industrial manufacturing & \ding{55} & - \\
      Fontanella et al. \cite{fontanella2024diffusion} & 2024 & IEEE TMI & Medical imaging (Brain) & \ding{51} & - \\
      FinDiff \cite{sattarov2023findiff} & 2023 & ACM ICAIF & Financial data & \ding{51} & - \\
      NetDiffus \cite{sivaroopan2024netdiffus} & 2024 & COMNET & Network traffic analysis & \ding{51} & - \\ \bottomrule
  \end{tabular}}}
\vspace{-15px}
\end{table}

\subsection{Performance Comparison}\label{sec8.3}

Leveraging the powerful capabilities of DMs, several frameworks have emerged as effective tools for generating synthetic anomalies with remarkable realism and diversity. As shown in \cref{tab:7key}, AnomalyDiffusion \cite{hu2024anomalydiffusion} demonstrates superior performance over traditional GAN-based methods like DefectGAN \cite{zhang2021defectgan} and DFMGAN \cite{duan2023fewshot}, achieving the highest Inception Score (1.8) and IC-LPIPS (0.32) on the MVTec AD dataset. Beyond AnomalyDiffusion, methods like CUT \cite{sun2024cut} further advance the field by leveraging Stable Diffusion for generating realistic anomalies without retraining, demonstrating universality across unseen data and novel anomaly types, which facilitates targeted anomaly creation for applications like data augmentation. Additionally, conditioning strategies play a crucial role in guiding the generation process. For example, SDAS in RealNet \cite{zhang2024realnet} employ a diffusion-based synthesis strategy specifically designed to mimic the distribution of real anomalies, creating a diverse set of synthetic examples for training robust anomaly detection models. Moreover, vision-language models \cite{dev2024reflipvad} can effectively leverage textual descriptions to condition the generation process, enabling fine-grained control over the characteristics of generated anomalies.
Since anomaly-generation studies evaluate realism, diversity, localization utility, or downstream detector gains under different protocols, the comparison reports the most relevant metric cues while the surrounding text explains the main limitation: cross-paper numbers are informative for orientation but not a controlled benchmark.

\begin{table}[t!]
  \centering
  \caption{Performance comparison of key AG models on the MVTec dataset using IS and IC-LPIPS metrics, with public implementations when available.}
  \label{tab:7key}
  \vspace{-0.1cm}
  \setlength{\tabcolsep}{2.4mm}{
  \resizebox{0.48\textwidth}{!}{
    \begin{tabular}{@{}lccccc@{}}
      \toprule
      \multirow{2}{*}{\textbf{Method}} & \multirow{2}{*}{\textbf{Year}} & \multirow{2}{*}{\textbf{Venue}} & \multicolumn{2}{c}{\textbf{Metric}} & \multirow{2}{*}{\textbf{Code}} \\ \cmidrule(l){4-5}
       &  &  & IS & IC-L &  \\ \midrule
      DefectGAN \cite{zhang2021defectgan} & 2021 & WACV & 1.69 & 0.15 & - \\
      SDGAN \cite{niu2020defect} & 2020 & IEEE TASE & 1.71 & 0.13 & - \\
      DiffAug \cite{zhao2020differentiable} & 2020 & NeurIPS & 1.58 & 0.09 & - \\
      Crop\&Paste \cite{lin2021fewshot} & 2021 & IEEE ICME & 1.51 & 0.14 & - \\
      DFMGAN \cite{duan2023fewshot} & 2023 & AAAI & 1.72 & 0.20 & - \\
      CDC \cite{ojha2021fewshot} & 2021 & CVPR & 1.65 & 0.07 & - \\
      AnomalyDiffusion \cite{hu2024anomalydiffusion} & 2024 & AAAI & 1.80 & 0.32 & \href{https://github.com/sjtuplayer/anomalydiffusion}{\textcolor{blue}{\faLink}} \\ \bottomrule
  \end{tabular}}}
\vspace{-15px}
\end{table}

\subsection{Applications}\label{sec8.4}

Anomaly generation with DMs has found diverse applications, notably in data augmentation \cite{hu2024anomalydiffusion}.  In scenarios with limited anomaly samples, synthetically generated anomalies can augment training datasets, consequently improving the performance and robustness of AD models, which is particularly valuable in areas like industrial inspection \cite{yao2024glad} where acquiring real-world anomalies is challenging.  Additionally, generating a wide range of potential anomalies facilitates model robustness testing \cite{bercea2023mask} and allows researchers to evaluate detection models under diverse conditions.  Through this process, vulnerabilities can be identified and the reliability of anomaly detection systems can be enhanced.  Anomaly generation also enables anomaly visualization \cite{pintilie2023time}, aiding in understanding anomalous event characteristics and interpreting model predictions. For example, in medical imaging, visualizing generated anomalies can assist clinicians in identifying subtle abnormalities.  Furthermore, some methods \cite{tebbe2023d3ad,tebbe2024dynamic} leverage dynamic noise addition during diffusion, guided by initial anomaly predictions, to improve anomaly localization, particularly for anomalies of varying scales.

\section{Model Evaluation}\label{sec9}
Evaluation opens with a cross-modality overview before the detailed metric and benchmark subsections: metrics should be interpreted as task-specific evidence for ranking, localization, or temporal consistency~\cite{muller2022guideline,sorbo2024navigating}, while benchmark datasets define the anomaly prevalence, annotation granularity, and domain shift conditions under which those metrics remain meaningful~\cite{han2022adbench}. In turn, the overview anchors the later modality-by-modality discussion in a common evaluation scaffold instead of treating each modality in isolation.
\subsection{Evaluation Metrics}\label{sec9.1}
\begin{figure}[t!]
  \centering
\includegraphics[width=0.48\textwidth]{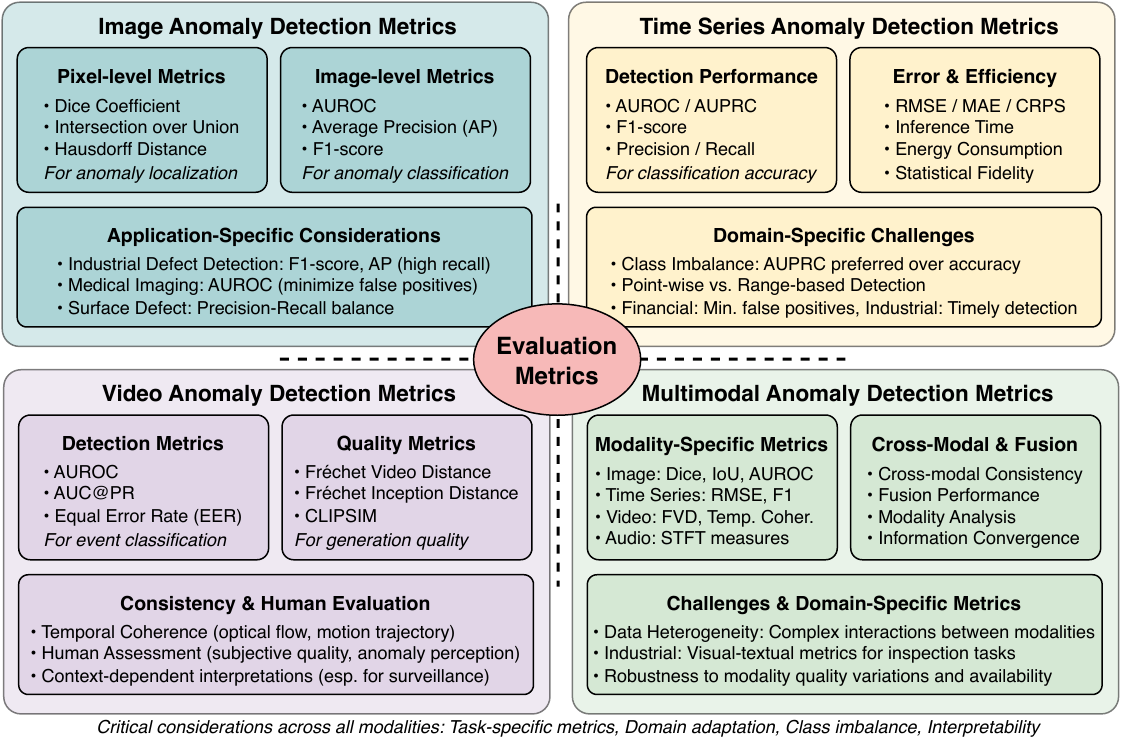}
\caption{Comprehensive evaluation metrics for ADGDM across different data modalities.}
  \label{fig:11}
  \vspace{-15px}
\end{figure}

Effective ADGDM relies on robust evaluation metrics tailored to specific data modalities. As shown in \cref{fig:11}, evaluation frameworks must address the distinct characteristics of each data type through specialized metrics.
Beyond listing metrics, a practical survey must also clarify when each metric is trustworthy. AUROC is useful for threshold-free ranking but can obscure deployment risk under severe class imbalance~\cite{muller2022guideline}; AUPRC and F1 better reflect rare-event retrieval but are threshold-sensitive~\cite{zhang2023diffusionad}; point-wise scores reward local detection whereas range-based metrics better capture contiguous temporal anomalies~\cite{zhang2020reconstruct}; localization metrics such as Dice/IoU matter when the actionability of an alert depends on spatial precision~\cite{yan2022impact}; and cross-modal consistency metrics are only meaningful when alignment quality and missing-modality behavior are reported alongside them~\cite{li2019deepa,liu2025multimodal}.

\subsubsection{Image Anomaly Detection Metrics}\label{sec9.1.1}

IAD employs evaluation metrics reflecting the task's multifaceted nature, categorized into pixel-level metrics for anomaly localization and image-level metrics for classification \cite{muller2022guideline}. Pixel-level localization relies on the Dice coefficient and Intersection over Union (IoU) to measure overlap between predicted and ground truth anomaly masks \cite{yan2022impact}, while the Hausdorff Distance quantifies maximum boundary distances, proving particularly valuable in medical imaging where precise localization is crucial \cite{hu2023anodode}. For classification tasks, AUROC measures the model's ability to distinguish between normal and anomalous images \cite{sheng2024surface}, while Average Precision addresses class imbalance inherent in anomaly datasets \cite{zhang2023diffusionad}, and F1-score balances precision and recall to reflect both correct identification and false positive avoidance \cite{tailanian2024uflow}. Metric selection depends on application requirements, with industrial defect detection prioritizing high recall through F1-score or AP, whereas medical applications favor AUROC to minimize false positives.

\subsubsection{Video Anomaly Detection Metrics}\label{sec9.1.2}
VAD evaluation requires assessing both detection accuracy and video quality while maintaining temporal consistency \cite{liu2024generalized}. Detection metrics including AUROC, AUC@PR, and Equal Error Rate quantify distinguishing capabilities between normal and anomalous events \cite{zaheer2024clustering}, with AUC@PR particularly relevant given class imbalance challenges. Video quality assessment employs Fréchet Video Distance, Fréchet Inception Distance, and CLIPSIM to evaluate generated video fidelity through feature distribution comparisons and semantic similarity measures \cite{rai2024video}. Temporal coherence metrics ensure smooth frame transitions by measuring optical flow consistency and motion trajectory smoothness \cite{tur2023exploring}, while human evaluation provides subjective assessment for subtle or context-dependent anomalies.

\subsubsection{Time Series Anomaly Detection Metrics}\label{sec9.1.3}
TSAD metrics encompass detection performance, error measurement, computational efficiency, and statistical fidelity \cite{sorbo2024navigating}. Detection capabilities utilize AUROC, AUPRC, F1-score, precision, and recall \cite{goswami2023unsupervised}, while error metrics like RMSE, MAE, and CRPS quantify prediction accuracy for forecasting-based methods \cite{wu2022developing}. Computational considerations including inference time and energy consumption prove critical for deployment \cite{park2024spectral}, alongside statistical fidelity metrics ensuring synthetic data reflects underlying distributions \cite{pintilie2023time}. Unique challenges include severe class imbalance where normal data vastly outnumbers anomalies \cite{zhao2022comparative}, distinctions between point-wise and range-based detection requiring different evaluation approaches \cite{zhang2020reconstruct}, and domain-specific requirements where financial applications prioritize false positive minimization while industrial monitoring emphasizes timely detection \cite{li2024difftad}.

\subsubsection{Multimodal Anomaly Detection Metrics}\label{sec9.1.4}
MAD evaluation addresses both modality-specific and cross-modal performance aspects through comprehensive metric frameworks \cite{wei2022msaf}. Individual modalities employ standard metrics like Dice coefficient, IoU, and Hausdorff Distance for localization alongside AUROC, AP, and F1-score for classification. Cross-modal assessment introduces consistency metrics quantifying information alignment between sources \cite{li2019deepa} and fusion performance metrics evaluating multimodal integration benefits over single-modality approaches \cite{shankar2021neural}. Application-specific adaptations combine visual-textual data for industrial inspection \cite{hu2024intermediate} or specialized audio-visual metrics for surveillance, while addressing challenges from data heterogeneity, lack of standardized benchmarks \cite{liang2023highmodality}, and the need for interpretable metrics capturing complementary information across modalities \cite{liu2025multimodal}.

\subsection{Benchmark Datasets}\label{sec9.2}

Evaluating AD requires access to well-established benchmark datasets spanning industrial inspection, medical screening, cyber-physical telemetry, and open-world video monitoring. To keep the main manuscript focused, we retain a concise cross-modality summary here and move the extended dataset-by-dataset comparison to the supplementary material, where \cref{supp:tab:supp-dataset-summary} consolidates representative metric cues, application domains, and practical bottlenecks.
Dataset choice also constrains metric interpretation: factory-inspection benchmarks reward localization fidelity~\cite{bergmann2019mvtec,zou2022spotdifference}, long-horizon telemetry benchmarks expose latency and range-detection trade-offs~\cite{su2019robust,hundman2018detecting}, and multimodal benchmarks remain limited by inconsistent annotation granularity and fusion protocols~\cite{liang2023highmodality}. For this reason, the ``best'' metric is scenario-dependent and should be read together with dataset bias, anomaly prevalence, and the intended deployment objective summarized below and expanded further in \cref{supp:tab:supp-dataset-summary}.

\begin{itemize}
\item \textbf{IAD benchmarks} such as \textit{MVTec AD}, \textit{MVTec 3D-AD}, \textit{LOCO}, \textit{VisA}, and \textit{BraTS} emphasize localization fidelity, small-defect sensitivity, and domain transfer between industrial and medical settings.
\item \textbf{TSAD benchmarks} including \textit{SMD}, \textit{SMAP}, \textit{MSL}, and \textit{SWaT} expose long-range dependency, point-vs-range anomaly scoring, and cyber-physical attack-detection trade-offs.
\item \textbf{VAD/TAD benchmarks} such as \textit{ShanghaiTech}, \textit{UCF-Crime}, \textit{CrossTask}, \textit{COIN}, \textit{TEP}, and \textit{ADBench} broaden coverage toward surveillance, procedural deviation, and structured-record anomaly settings, with the supplementary table organizing their metric cues and practical bottlenecks in a unified view.
\end{itemize}

\subsubsection{Image Anomaly Detection Datasets}\label{sec9.2.1}

\textit{MVTec AD} \cite{bergmann2019mvtec} serves as the de facto standard for industrial anomaly detection, encompassing 15 object categories with diverse defect types.  Its extension, \textit{MVTec 3D-AD} \cite{bergmann2022mvtec}, incorporates depth information, facilitating research on 3D anomaly detection.  Additionally, \textit{MVTec LOCO} \cite{bergmann2022dents} focuses on logical anomalies characterized by structural or compositional irregularities.  Other industrial datasets include \textit{BTAD} \cite{mishra2021vtadl}, \textit{VisA} \cite{zou2022spotdifference}, \textit{KSDD2} \cite{bozic2021mixed}, \textit{MPDD} \cite{jezek2021deep}, and \textit{PCB-Bank} \cite{yao2025glad}, each presenting unique characteristics for evaluating different aspects of diffusion model performance.  Finally, the medical dataset \textit{BraTS} \cite{kazerooni2024brain} provides MRI scans for brain tumor segmentation, introducing challenges related to subtle anomalies and the critical need for accurate detection in medical applications.

\subsubsection{Video Anomaly Detection Datasets}\label{sec9.2.2}
Industrial surveillance datasets including \textit{UCSD Ped2} \cite{mahadevan2010anomaly} and \textit{CUHK Avenue} \cite{lu2013abnormal} provide foundational benchmarks for non-pedestrian detection and unusual pedestrian behaviors, while comprehensive datasets \textit{ShanghaiTech} \cite{sultani2018realworld} and \textit{UCF-Crime} \cite{sultani2018realworld} offer diverse evaluation scenarios across 13 scenes and crime categories respectively. The most challenging benchmark, \textit{NWPU Campus} \cite{cao2023new,cao2025scenedependent}, features 43 scenes with scene-dependent anomaly definitions where identical events may be normal or abnormal based on context. Complementing surveillance data, \textit{UBnormal} \cite{acsintoae2022ubnormal} provides densely annotated videos for pixel-level localization, while instructional datasets \textit{CrossTask} \cite{zhukov2019crosstask} and \textit{COIN} \cite{tang2019coin} evaluate procedural anomaly detection through deviations from standard procedures.

\subsubsection{Time Series Anomaly Detection Datasets}\label{sec9.2.3}
Industrial monitoring drives TSAD benchmark development, with \textit{SMD} \cite{su2019robust} providing server metrics for multivariate dependency evaluation, while spacecraft datasets \textit{SMAP} \cite{hundman2018detecting} and \textit{MSL} \cite{hundman2018detecting} offer real-world telemetry scenarios. Moreover, industrial control contributions include \textit{SWaT} \cite{goh2017dataset} featuring real-world water purification testbed data with simulated attacks, specialized datasets \textit{TEP} \cite{chiang2000fault} with chemical process environments, \textit{Batchbenchmark} \cite{vanimpe2015extensive} for manufacturing processes, and comprehensive \textit{ADBench} \cite{han2022adbench} containing 57 real-world datasets across domains. The diversity in temporal characteristics and anomaly prevalence provides comprehensive evaluation settings for diffusion model performance in capturing temporal correlations essential for robust TSAD applications.

\section{Discussion}\label{sec10}
\subsection{Open Challenges}\label{sec10.1}
The application of ADGDM has shown promising results, yet several critical challenges remain to be addressed to achieve robust, scalable, and generalizable solutions.

\noindent\textbf{Scalability.} Scaling DMs to high-dimensional data, such as high-resolution images or long video sequences, remains challenging due to significant memory and computational demands~\cite{chen2023speed}. Iterative denoising processes exacerbate scalability issues, limiting applicability in resource-intensive settings~\cite{golnari2023loraenhanced}. Developing efficient architectures, such as latent diffusion or optimized sampling techniques, is crucial to handlling complex multimodal inputs effectively while maintaining detection accuracy~\cite{li2024unimog}.

\noindent\textbf{Generalization.} Achieving robust generalization for diffusion models across domains, e.g., from industrial to medical imaging, is hindered by domain shifts and varying anomaly types~\cite{carvalho2023invariant,jin2024survey}. Models trained on specific datasets often fail to adapt to new contexts, reducing their practical utility~\cite{fernando2022deep,chandrakala2023anomaly}. Transfer learning and domain-adaptive techniques are essential to enhance cross-domain performance, ensuring DMs can handle diverse data distributions and anomaly characteristics effectively~\cite{hu2023anodode}.

\noindent\textbf{Class Imbalance.} Detecting rare anomalies in imbalanced datasets, where normal samples dominate, poses a significant challenge for diffusion models~\cite{muller2022guideline}. Overfitting to normal data distributions reduces sensitivity to subtle anomalies, compromising detection accuracy~\cite{celaya2024generalized}. Techniques like anomaly augmentation or robust scoring mechanisms are needed to improve model performance on rare events, ensuring reliable detection in critical applications such as industrial monitoring~\cite{jin2024dualanodiff}.

\noindent\textbf{Computational Efficiency.} High computational cost of DMs during inference, driven by iterative denoising steps, hinders deployment in real-time or resource-constrained environments~\cite{farid2023latent}, thereby limiting their practicality for applications like industrial monitoring~\cite{defard2021padim}. Across the literature, the most consistent efficiency cues are whether computation occurs in pixel space or latent space, whether sampling uses the full reverse chain or an accelerated solver, and whether additional distillation/parallelization is needed before deployment. \Cref{tab:efficiency-compare} summarizes representative training-side and inference-side efficiency cues across modalities so that these trade-offs can be compared in one place. For example, latent-space designs reduce the per-step state size for medical or multimodal volumes~\cite{pinaya2022fast,li2024unimog}, DDIM-style or shortened reverse paths trade some fidelity for faster image reconstruction~\cite{fontanella2024diffusion,wyatt2022anoddpm}, progressive distillation compresses TSAD sampling depth at the cost of extra training stages~\cite{tian2024prodiffad}, and hybrid tabular generators exploit parallel synthesis to reduce wall-clock latency but introduce front-end encoding overhead~\cite{suh2024timeautodiff}. Consequently, techniques such as model distillation, fewer-step sampling, or hardware-aware optimizations are essential to reduce inference latency while preserving AD accuracy~\cite{roth2022total}.

\begin{table}[t!]
\centering
\caption{Representative efficiency cues across ADGDM.}
\label{tab:efficiency-compare}
\vspace{-0.1cm}
\setlength{\tabcolsep}{5mm}{
\resizebox{0.48\textwidth}{!}{
\begin{tabular}{@{}p{1.45cm}p{0.5cm}p{1.55cm}p{1.55cm}p{2.2cm}@{}}
\toprule
\textbf{Method} & \textbf{Mod.} & \textbf{Train. Cue} & \textbf{Infer. Cue} & \textbf{Trade-off} \\ \midrule
Latent DDPM \cite{pinaya2022fast} & IAD & Latent 3D modeling & Smaller latent state & Faster sampl. may miss tiny lesions \\
DDPM-DDIM \cite{fontanella2024diffusion} & IAD & Standard diffusion training & Shortened reverse path & Lower latency vs. recon. loss \\
ProDiffAD \cite{tian2024prodiffad} & TSAD & Progressive distillation stage & Reduced sampling depth & Better speed vs. calibration issues \\
NPSR \cite{lai2023nominality} & TAD & Hybrid encoder-diffusion & Parallel table synthesis & Faster speed vs. extra encoding cost \\
UNIMO-G \cite{li2024unimog} & MAD & Large multimodal backbone & Cross-modal condition & Stronger context vs. memory overhead \\
\bottomrule
\end{tabular}}}
\vspace{-12px}
\end{table}

\noindent\textbf{Robustness.} DMs are vulnerable to adversarial perturbations and noisy inputs, which can distort anomaly scoring and compromise reliability~\cite{lu2023removing}, particularly in domains like medical diagnostics, where robustness is paramount~\cite{li2023fast}. Given these vulnerabilities, developing adversarial training strategies or noise-robust scoring mechanisms is necessary to enhance model stability, ensuring consistent performance under challenging conditions and safeguarding against malicious or erroneous inputs~\cite{lin2025adversarial}.

\noindent\textbf{Explainability.} Although many ADGDM pipelines yield accurate anomaly scores, critical applications require explanations that identify what changed, where the evidence is located, and why the model is confident. Reconstruction-difference attribution and saliency maps help localize suspicious regions or timestamps, but they can still over-highlight artifacts introduced by the denoiser itself~\cite{zhang2024interpretable}. Counterfactual anomaly generation offers a complementary route by editing the same sample toward nearby normal or anomalous states, while text-assisted and human-in-the-loop verification frameworks are increasingly needed in medical screening, industrial inspection, and cyber defense to determine whether a highlighted deviation is actionable or merely distribution drift~\cite{lengerich2023llms,chen2023interpretation,zhuo2019explainable}.

\noindent\textbf{Deployment-Oriented Case Studies.} In industrial inspection, diffusion models are typically paired with high-resolution cameras and strict cycle-time budgets, so the practical question is whether a method can localize a defect fast enough for inline rejection while still providing defect-type evidence to quality engineers~\cite{defard2021padim,yao2024glad}. In medical screening, the same reconstruction machinery must instead prioritize lesion-level sensitivity, uncertainty communication, and clinician-auditable rationales because a visually plausible reconstruction can still suppress subtle but clinically relevant anomalies~\cite{pinaya2022fast,li2023fast}.
A complementary deployment pattern appears in cyber-physical and network monitoring, where multivariate telemetry, alarm logs, and topology context arrive asynchronously~\cite{kheddar2025reinforcement,goh2017dataset,su2019robust}. Here, diffusion-based detectors are most useful when they summarize which channels drifted first, whether the event resembles attack-like contamination or routine operating-mode changes, and how much latency the reverse process introduces before analysts can triage the alert. Together, the examples reinforce that deployment success depends jointly on modality fit, explanation format, and efficiency constraints rather than on accuracy alone.

\subsection{Future Opportunities}\label{sec10.2}
ADGDM presents several promising opportunities for future research and development, with prospects to enhance efficiency, integrate novel conditioning, combine with other techniques, explore new modalities, and leverage LLMs for intelligent detection, that could significantly advance the capabilities and practical applications of this field.

\noindent\textbf{Efficient DM architectures.} High computational costs limit DM deployment in real-time anomaly detection applications, particularly for high-resolution images and videos. Dynamic step size computation~\cite{tebbe2023d3ad} addresses this through initial anomaly prediction guidance, while latent space projections and quantization techniques reduce memory demands. Moreover, alternatives to full-length Markov chain diffusion~\cite{wyatt2022anoddpm} and multi-scale simplex noise diffusion offer pathways to scaling DMs to high-resolution imagery without compromising detection quality.

\noindent\textbf{New Conditioning Strategies.} Enhanced conditioning strategies incorporate prior knowledge and contextual information to improve anomaly detection performance~\cite{farid2023latent}. Multimodal conditioning through text or audio provides complementary detection cues~\cite{hu2023cognitively}, while domain-specific knowledge integration improves detection specificity. Temporal conditioning benefits time-series and video analysis~\cite{pintilie2023time}, while feedback loops using anomaly scores~\cite{tebbe2024dynamic,tebbe2023d3ad} enhance localization accuracy. Adaptive conditioning strategies that respond to input characteristics offer promising solutions for heterogeneous datasets.

\noindent\textbf{DMs with Other AD techniques.} Hybrid approaches combining DMs with existing methods yield superior performance through complementary strengths. Clustering integration leverages generative capabilities for representative normal data synthesis~\cite{yao2024glad}, while one-class classifiers benefit from DM-generated samples~\cite{acsintoae2022ubnormal}. Notable methods include ODD~\cite{wang2023odd} combining similarity networks with outlier exposure, and dual conditioning frameworks for multi-class detection through category-specific reconstructions~\cite{zhan2024enhancing}.

\noindent\textbf{New Modalities and Applications.} Current ADGDM focuses primarily on images, videos, and time series, leaving opportunities in broader modalities and applications. Tabular data applications in finance and healthcare show promise despite mixed data type challenges~\cite{livernoche2023diffusion}. Multimodal detection~\cite{tebbe2024dynamic} and extensions to audio or 3D point clouds could enable new applications, while time series advances address irregular sampling and long-term dependencies~\cite{pintilie2023time}, combined with self-supervised learning~\cite{li2024selfsupervised}.

\noindent\textbf{AD with LLMs and VLMs.} LLMs and VLMs can enhance diffusion-based detection, but their most promising role is as a coordination and explanation layer over diffusion outputs rather than as a replacement for modality-specific detectors. A practical framework space includes knowledge-grounded text conditioning before diffusion, post-hoc explanation generation from reconstruction maps and anomaly scores~\cite{lengerich2023llms}, cross-modal alignment of image/text/audio evidence in multimodal settings~\cite{chen2024explore,azarafza2024hybrida}, long-horizon temporal summarization for TSAD~\cite{zhang2024largeb}, and agent-assisted monitoring loops that convert synthetic anomaly generation into scenario-driven augmentation or triage checklists~\cite{henrique2023stochastic,qi2023large}. We emphasize that these are literature-grounded integration frameworks rather than new empirical claims, and their value is highest in medical diagnosis and cybersecurity where analysts need grounded textual reasoning, policy-aware triage, and human-verifiable recommendations instead of a single black-box anomaly score~\cite{shao2024quantifying,barua2024exploring}.

\section{Conclusion}\label{sec11}
In this survey, we comprehensively examine anomaly detection and generation with diffusion models (ADGDM) across diverse data modalities. We explore various conditioning strategies and their impact on performance while highlighting adaptations for image, video, time series, and tabular data. We also explore emerging multimodal approaches and assess the potential of DMs for synthetic anomaly generation. Researchers currently face challenges including scalability, generalization, class imbalance, computational efficiency, and robustness across applications. Future research opportunities encompass developing efficient architectures, novel conditioning strategies, hybrid approaches integrating traditional techniques, and leveraging large language models.

\ifCLASSOPTIONcaptionsoff
  \newpage
\fi

{\small
\bibliographystyle{IEEEtran}
\begin{spacing}{0.94} 
\bibliography{refs}
\end{spacing}
}

\end{document}